\documentclass{article}

\usepackage[final,nonatbib]{neurips_2021}

\usepackage[numbers,sort&compress]{natbib} 

\usepackage[utf8]{inputenc} 
\usepackage[T1]{fontenc}    
\usepackage{booktabs}       
\usepackage{amsfonts}       
\usepackage{nicefrac}       
\usepackage{microtype}      
\usepackage{xcolor}         

\usepackage{multicol}
\usepackage{multirow}
\usepackage{xspace}
\usepackage{subcaption}
\usepackage{paralist}
\usepackage{mathtools}
\usepackage{bbm}
\usepackage{wrapfig}
\usepackage[hidelinks]{hyperref}    
\usepackage{url}
\newcommand{\vspacecaption}{\vspace{-1pt}}
\newcommand{\vspacecaptionlow}{\vspace{-1pt}}
\newcommand{\vspacesubcaption}{\vspace{-1pt}}

\newcommand{\vspacefigurecaptiontop}{\vspace{-3pt}}
\newcommand{\vspacefigurecaption}{\vspace{-3pt}}
\newcommand{\vspacefiguretopofpage}{\vspace{-21pt}} 
\newcommand{\vspacefigurebetweenrows}{\vspace{0pt}}

\usepackage{algorithm}
\usepackage[noend]{algpseudocode}

\usepackage{threeparttable}

\usepackage{tikz}
\usetikzlibrary{bayesnet}

\allowdisplaybreaks

\usepackage{amsthm}
\usepackage{adjustbox}

\usepackage{Definitions}

\newcommand\ie{\text{i.e.}\xspace}
\newcommand\eg{\text{e.g.}\xspace}

\newcommand\pa{\text{pa}}

\newcommand\mcmcmc{\text{MC${}^3$}\xspace}

\newcommand\erdosrenyi{\text{Erd{\H{o}}s-R{\'e}nyi}\xspace}

\title{DiBS: Differentiable Bayesian Structure Learning}
             
\author{
  Lars Lorch\\
  ETH Zurich\\
  Zurich, Switzerland\\
  \texttt{lars.lorch@inf.ethz.ch}\\
  \And
  Jonas Rothfuss\\
  ETH Zurich\\
  Zurich, Switzerland\\
  \texttt{jonas.rothfuss@inf.ethz.ch}\\
  \AND 
  Bernhard Sch{\"o}lkopf\\
  MPI for Intelligent Systems\\
  T{\"u}bingen, Germany\\
  \texttt{bs@tuebingen.mpg.de}\\
  \And
  Andreas Krause\\
  ETH Zurich\\
  Zurich, Switzerland\\
  \texttt{krausea@ethz.ch}
}

\begin{document}

\maketitle

\begin{abstract}
Bayesian structure learning allows inferring Bayesian network structure from data while reasoning about the epistemic uncertainty---a key element towards enabling active causal discovery and designing interventions in real world systems. In this work, we propose a general, fully {\em differentiable} framework for {\em Bayesian structure learning (DiBS)} that operates in the continuous space of a latent probabilistic graph representation. Contrary to existing work, DiBS is agnostic to the form of the local conditional distributions and allows for joint posterior inference of both the graph structure and the conditional distribution parameters. This makes our formulation directly applicable to posterior inference of complex Bayesian network models, e.g., with nonlinear dependencies encoded by neural networks. 
Using DiBS, we devise an efficient, general purpose variational inference method for approximating distributions over structural models. 
In evaluations on simulated and real-world data, our method significantly outperforms related approaches to joint posterior inference.\footnote{Our Python \texttt{JAX} implementation of DiBS is available at:  
\href{https://github.com/larslorch/dibs}{\color{blue}\texttt{https://github.com/larslorch/dibs}}
} 
\end{abstract}

\vspacecaption
\section{Introduction}\label{sec:introduction}
\vspacecaptionlow

Discovering the statistical and causal dependencies that underlie the variables of a data-generating system is of central scientific interest.
Bayesian networks (BNs) \citep{pearl1988probabilistic} and structural equation models are commonly used for this purpose~\citep{pe2001inferring,sachs2005causal,van2006application,weber2012overview}.
Structure learning, the task of learning a BN from observations of its variables, is well-studied, but computationally very challenging due to the combinatorially large number of candidate graphs and the constraint of graph acyclicity.

While structure learning methods arrive at a single plausible graph or its Markov equivalence class (MEC), \eg, \citep{chickering2003optimal,spirtes2000causation,tsamardinos2006max,zheng2018dags}, \emph{Bayesian structure learning} aims to infer a full posterior distribution over BNs given the observations.
A distribution over structures allows quantifying the epistemic uncertainty and the degree of confidence in any given BN model, \eg, when the amount of data is small.
Most importantly, downstream tasks such as experimental design and active causal discovery rely on a posterior distribution over BNs to quantify the information gain from specific interventions and uncover the causal structure in a small number of experiments \citep{murphy2001active,tong2001active,cho2016reconstructing,ness2017bayesian,agrawal2019abcdstrategy,von2019optimal}.

\looseness - 1 A key challenge in Bayesian structure learning is working with a posterior over BNs---a distribution over the joint space of (discrete) directed acyclic graphs and (continuous) conditional distribution parameters. 
Most of the practically viable approaches to Bayesian structure learning revolve around Markov chain Monte Carlo (MCMC) sampling in combinatorial spaces and bootstrapping of classical score and constraint-based structure learning methods, \eg, in causal discovery \citep{murphy2001active,tong2001active,cho2016reconstructing,ness2017bayesian,agrawal2019abcdstrategy,von2019optimal}.
However, these methods marginalize out the parameters and thus require a closed form for the marginal likelihood of the observations given the graph to remain tractable. 
This limits inference to simple and by now well-studied linear Gaussian and categorical BN models \citep{geiger1994learning,geiger2002parameter,heckerman1995learning} and makes it difficult to infer more expressive BNs that, \eg, model nonlinear relationships among the variables.
Due to the discrete nature of these approaches, recent advances in approximate inference and gradient-based optimization could not yet be translated into similar performance improvements in Bayesian structure learning.

\looseness - 1
In this work, we propose a novel, {\em fully differentiable framework for Bayesian structure learning} (DiBS) that operates in the continuous space of a latent probabilistic graph representation. 
Contrary to existing work, our formulation is agnostic to the distributional form of the BN
and allows for inference of the joint posterior over both the conditional distribution parameters and the graph structure.
This makes our approach directly applicable to more flexible BN models where neither the marginal likelihood nor the maximum likelihood parameter estimate have a closed form.
We instantiate DiBS with the particle variational inference method of \citet{liu2016stein} and present a general purpose method for approximate Bayesian structure learning.
In our experiments on synthetic and real-world data, DiBS outperforms all alternative approaches to joint posterior inference of graphs and parameters and when modeling nonlinear interactions among the variables, often by a significant margin.
This allows us to narrow down plausible causal graphs with greater precision and make better predictions under interventions---an important stepping stone towards active causal discovery.

\vspacecaption
\section{Background}\label{sec:background}
\vspacecaptionlow

\textbf{Bayesian networks\quad}
A Bayesian network $(\Gb, \Thetab)$ models the joint density $p(\xb)$ of a set of  $d$ variables $\xb = x_{1:d}$ using (1) a directed acyclic graph (DAG) $\Gb$ encoding the conditional independencies of $\xb$ and (2) parameters $\Thetab$ defining the local conditional distributions of each variable given its parents in the DAG.
When modeling $p(\xb)$ using a BN, each variable is assumed to be independent of its non-descendants given its parents, thus allowing for a compact factorization of the joint $p(\xb \given \Gb, \Thetab)$ into a product of local conditional distributions for each variable and its parents in $\Gb$.

\textbf{Bayesian inference of BNs\quad}
Given independent observations $\Dcal = \{\xb^{(1)}, \dots, \xb^{(N)}\}$, we consider the task of inferring a \emph{full posterior} density over Bayesian networks that model the observations.
Following \citet{friedman2003being}, given a prior distribution over DAGs $p(\Gb)$ and a prior over BN parameters $p(\Thetab \given \Gb)$, Bayes' Theorem yields the joint and marginal posterior distributions
\begin{align}
	p(\Gb, \Thetab \given \Dcal) 
		&\propto p(\Gb) p(\Thetab \given \Gb) p(\Dcal \given \Gb, \Thetab) \label{eq:bayesian-structure-learning-joint-posterior}\;,\\
	p(\Gb \given \Dcal) 
		&\propto p(\Gb) p(\Dcal \given \Gb) \label{eq:bayesian-structure-learning-marg-posterior} 
\end{align}
where $p(\Dcal \given \Gb)$~$=$~$\int p(\Thetab \given \Gb) p(\Dcal \given \Gb, \Thetab) d\Thetab$ is the marginal likelihood. Thus, $p(\Gb \given \Dcal)$ in (\ref{eq:bayesian-structure-learning-marg-posterior}) is only tractable in special conjugate cases where the integral over $\Thetab$ can be computed in closed form.
The Bayesian formalism allows us to compute expectations of the form
\begin{align} \label{eq:bayesian-structure-learning-goal}
	\EE_{p(\Gb, \Thetab \given \Dcal)} 
	\Big [
		f(\Gb, \Thetab)
	\Big ] 
	\quad \quad \text{or} \quad \quad
	\EE_{p(\Gb \given \Dcal)} 
	\Big [
		f(\Gb)
	\Big ] 
\end{align}
\looseness - 1 for any function $f$ of interest. 
For instance, to perform Bayesian model averaging, we would use $f(\Gb, \Thetab)$~$=$~$p(\xb \given \Gb, \Thetab)$ or $f(\Gb)$~$=$~$p(\xb \given \Gb)$, respectively~\citep{madigan1994model,madigan1995eliciting}.
In active learning of causal BN structures, a commonly used $f$
\vspace*{-3pt} 
is the expected decrease in entropy of $\Gb$ after an intervention
\citep{tong2001active,murphy2001active,cho2016reconstructing,agrawal2019abcdstrategy,von2019optimal}. 
Inferring either posterior is computationally challenging
because there are $\mathcal{O}(d! 2^{\binom{d}{2}})$ possible DAGs with $d$ nodes \citep{robinson1973counting}.
Thus, computing the normalization constant $p(\Dcal)$ is generally intractable. 

\textbf{Continuous characterization of acyclic graphs\quad} 
Orthogonal to the work on Bayesian inference, \citet{zheng2018dags} have recently proposed a differentiable characterization of acyclic graphs for structure learning.
In this work, we adopt the formulation of \citet{yu2019daggnn}, who show that a graph with adjacency matrix
$\Gb \in \{0,1\}^{d \times d}$ does not have any cycles if and only if $h(\Gb) = 0$, where
\begin{align}\label{eq:trace-exponential}
	h(\Gb) := \tr \Big [ (\Ib + \tfrac{1}{d} \Gb )^d \Big ] - d \;.
\end{align}
\looseness -1  If $h(\Gb) > 0$, the function can be interpreted as quantifying the \emph{cyclicity} or \emph{non-DAG-ness} of $\Gb$.
Follow-up work has leveraged this insight to model nonlinear relationships \citep{yu2019daggnn,lachapelle2019gradientbased,ng2019graph,zheng2020learning},
time-series data \citep{pamfil2020dynotears},
in the context of generative modeling \citep{yang2020causalvae,zhang2019dvae},
for causal inference \citep{ke2019learning,ng2019masked,brouillard2020differentiable},
and contributed to its theoretical understanding \citep{wei2020dags,ng2020convergence}. So far, a connection to Bayesian structure learning has been missing.

\vspacecaption
\section{Related Work}\label{sec:related-work}
\vspacecaptionlow
\looseness - 1 The existing literature on Bayesian Structure Learning predominantly focuses on inferring the marginal graph posterior $p(\Gb \given \Dcal)$.
Since this requires $p(\Dcal \given \Gb)$ to be tractable, inference is limited to BNs with linear Gaussian or Categorical conditional distributions~\citep{geiger1994learning,geiger2002parameter,heckerman1995learning}. 
By contrast, the formulation we introduce overcomes this fundamental restriction by allowing for joint inference of the graph and the parameters, thereby facilitating the active (causal) discovery of more expressive BNs.

\textbf{MCMC\quad} 
Sampling from the posterior over graphs is the most general approach to approximate Bayesian structure learning.
Structure MCMC (\mcmcmc) \citep{madigan1995bayesian,giudici2003improving} performs Metropolis-Hastings in the space of DAGs by changing one edge at a time. 
Several works try to remedy its poor mixing behavior \citep{eaton2007bayesian,grzegorczyk2008improving,kuipers2017partition}.
Alternatively, order MCMC draws samples in the smaller but still exponential space of node orders, 
which typically requires a hard limit on the maximum parent set size \citep{friedman2003being}.
Attempts to correct for its unintended structural bias are themselves NP-hard to compute and/or limit the parent size \citep{eaton2007bayesian,ellis2008learning}.
By performing variational inference in a continuous latent space, the method we propose circumvents such mixing issues and parent size limitations.

\textbf{Bootstrapping\quad}
The nonparametric DAG bootstrap \citep{friedmann1999bootstrap} performs model averaging by bootstrapping $\Dcal$, where each resampled data set is used to learn a single graph, \eg, using the GES or PC algorithms \citep{chickering2003optimal,spirtes2000causation}.
The obtained set of DAGs approximates the posterior by weighting each unique graph by its unnormalized posterior probability.
In simple cases, a closed-form maximum likelihood parameter estimate may be used to approximate the joint posterior~\citep{agrawal2019abcdstrategy},
but only if $p(\Dcal \given \Gb)$ is tractable in the first place.

\textbf{Exact methods\quad}
A few notable exceptions use dynamic programming to achieve exact marginal inference in time $O(d2^d)$, which is only feasible for $d$~$\leq$~$20$ nodes~\citep{koivisto2004exact,koivisto2006advances}.
In special cases, \eg, for tree structures or known node orderings, exact inference can be performed more efficiently~\citep{meila2013tractable,dash2004model}.

\vspacecaption
\section{A Fully Differentiable Framework for Bayesian Structure Learning}\label{sec:continuous-bayesian-structure-learning}
\vspacecaptionlow
%
\begin{wrapfigure}{r}{0.32\textwidth} 
\centering \vspace{-12pt}
\begin{tikzpicture}[x=1.0cm,y=0.5cm]
  \node[latent] (z) {$\Zb$};
  \node[latent, right=of z] (g) {$\Gb$};
  \node[latent, right=of g]  (theta) {$\Thetab$};
  \node[obs, below=of g, xshift=0.9cm] (x) {$\xb$};
  

  \edge {z} {g} ; %
  \edge {g} {theta} ; %
  \edge {g, theta} {x} ; %


  \plate {} {(x)} {$N$} ;

\end{tikzpicture}
\caption{ \looseness -1
Generative model of BNs with latent variable~$\Zb$. This formulation generalizes the standard Bayesian setup in~(\ref{eq:bayesian-structure-learning-joint-posterior}) where only $\Gb$, $\Thetab$, and $\xb$ are modeled explicitly. 
}
\vspace{-5pt}
\label{fig:cls-generative-model}
\end{wrapfigure}
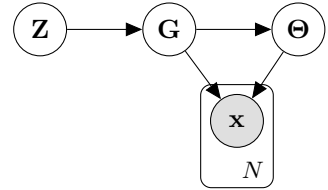
\vspacesubcaption
\subsection{General Approach}
\vspacesubcaption
With the goal of moving beyond the restrictive conjugate setups required by most discrete sampling methods for Bayesian structure learning, we propose to transfer the posterior inference task into the latent space of a probabilistic graph representation.
Our resulting framework is \emph{consistent} with the original Bayesian structure learning task in (\ref{eq:bayesian-structure-learning-goal}), enforces the \emph{acyclicity} of $\Gb$ via the latent space prior, and provides the \emph{score} of the continuous latent posterior, thus making general purpose inference methods applicable off-the-shelf.

Without loss of generality, we assume that there exists a latent variable $\Zb$ that models the generative process of $\Gb$.
Specifically, the default generative model described in Section \ref{sec:background} is generalized into
the following factorization:
\begin{align} \label{eq:own-generative}
	p(\Zb, \Gb, \Thetab, \Dcal) = p(\Zb) p(\Gb \given \Zb) p(\Thetab \given \Gb)  p(\Dcal \given \Gb, \Thetab)
\end{align}
Figure \ref{fig:cls-generative-model} displays the corresponding graphical model. 
The following insight provides us with an equivalence between the expectation we ultimately want to approximate and an expectation over the posterior of the continuous variable $\Zb$:

\begin{proposition}[Latent posterior expectation]\label{lemma:1}
	Under the generative model in (\ref{eq:own-generative}), it holds that
\begin{itemize}
	\item[\text{(a)}]~
	$\displaystyle
	\EE_{p(\Gb \given \Dcal)}  \Big [f(\Gb)\Big ] 
	~=~
	\EE_{p(\Zb \given \Dcal)} \Bigg [  
	\frac
		{\EE_{p(\Gb \given \Zb)}~ \big [ f(\Gb)p(\Dcal \given \Gb) \big ] }
		{\EE_{p(\Gb \given \Zb)}~ \big [       p(\Dcal \given \Gb) \big ] }
	\Bigg ]$ ~, ~~ \text{and}

	\item[\text{(b)}]~
	$\displaystyle
	\EE_{p(\Gb, \Thetab \given \Dcal)}  \Big [f(\Gb, \Thetab)\Big ] 
	~=~
	\EE_{p(\Zb, \Thetab \given \Dcal)} \Bigg [  
	\frac
		{\EE_{p(\Gb \given \Zb)}~ \big [ f(\Gb, \Thetab)p(\Thetab \given \Gb)p(\Dcal \given \Gb, \Thetab) \big ] }
		{\EE_{p(\Gb \given \Zb)}~ \big [ p(\Thetab \given \Gb)p(\Dcal \given \Gb, \Thetab) \big ] }
	\Bigg ]$ \;.
	
\end{itemize}
\end{proposition}
\looseness -1
A proof is provided in Appendix \ref{app:proof-lemma-1}.
Rather than approximating $p(\Gb \given \Dcal)$ or $p(\Gb, \Thetab \given \Dcal)$, our goal will be to infer $p(\Zb \given \Dcal)$ or $p(\Zb, \Thetab \given \Dcal)$ instead, which by Proposition \ref{lemma:1} allows us to compute expectations of the form in (\ref{eq:bayesian-structure-learning-goal}).
In the following, we first discuss how to define the two factors $p(\Gb \given \Zb)$ and $p(\Zb)$ in a way that models only directed acyclic graphs. Then, we provide the details necessary to perform black box inference of the posteriors of the continuous latent variable $\Zb$. 


\vspacesubcaption
\subsection{Representing DAGs in a Continuous Latent Space} \label{sec:graph-latent-space}
\vspacesubcaption

\textbf{Generative model of directed graphs\quad}
We define the latent variable $\Zb$ as consisting of two embedding matrices $\Ub,\Vb$~$\in$~$\RR^{k \times d}$, \ie, $\Zb$~$=$~$[\Ub, \Vb]$. 
Building on \citet{kipf2016variational}, we propose to use a bilinear generative model for the adjacency matrix $\Gb$~$\in$~$\{0,1\}^{d\times d}$ of directed graphs using the inner product between the latent variables in $\Zb$:
\begin{align}\label{eq:graph-latent-model}
	\hspace*{-8pt}p_\alpha(\Gb \given \Zb) &= 
	\prod_{i=1}^d
	\prod_{j \neq i}^d
	p_\alpha(g_{ij} \given \ub_i, \vb_j)
	~~~
	\text{with}
	~~~
	p_\alpha(g_{ij} = 1 \given \ub_i, \vb_j) = \sigma_\alpha(\ub^\top_i \vb_j)
\end{align}
where $\sigma_\alpha(x) = 1/(1 + \exp(-\alpha x))$ denotes the sigmoid function with inverse temperature $\alpha$.
We denote the corresponding matrix of edge probabilities in $\Gb$ given $\Zb$ by $\Gb_\alpha(\Zb) \in [0, 1]^{d \times d}$ with
\begin{align}\label{eq:graph-generative-model-matrix}
	\Gb_\alpha(\Zb)_{ij} := p_\alpha(g_{ij} = 1 \given \ub_i, \vb_j) \;.
\end{align}
Since we model acyclic graphs, which do not contain self-loops,  we set $p_\alpha(g_{ii} = 1 \given \Zb)$~$:=$~$0$.
The latent dimensionality $k$ trades off the complexity of the variable interactions with tractability during inference. For $k$~$\ge$~$d$, the matrix of edge probabilities $\Gb_\alpha(\Zb)$ is not constrained in rank, and the generative model in \eq{eq:graph-latent-model} can represent any adjacency matrix without self-loops. That said, the size of the latent representation $\Zb$ only grows $\mathcal{O}(d\hspace{-1pt}\cdot\hspace{-1pt}k)$ and, in principle, $k$ can be chosen independently of $d$. Note that the formulation in \eq{eq:graph-latent-model} models \emph{directed} graphs since
$\sigma_\alpha(\ub^\top_i \vb_j) \neq \sigma_\alpha(\ub^\top_j \vb_i)$.
We can even interpret $\ub_i$ and $\vb_i$ as node embeddings that may encode more information than mere edge probabilities, \eg, for graph neural networks. 
The fact that $p_\alpha(\Gb \given \Zb)$ is invariant to orthogonal transformations of $\Ub$ and $\Vb$ is not an issue in practice.


\textbf{Acyclicity via the latent prior distribution\quad}
\looseness -1 A major constraint in learning BNs is the acyclicity of $\Gb$. 
With a latent graph model $p(\Gb \given \Zb)$ in place, we design the prior $p(\Zb)$ to act as a soft constraint enforcing that only DAGs are modeled.
Specifically, we define the prior of $\Zb$ as the product of independent Gaussians with a Gibbs distribution that penalizes the \emph{expected cyclicity} of $\Gb$ given $\Zb$:
\begin{align}
	p_\beta(\Zb) &\propto \exp 
	\left ( 
		- \beta ~
		\EE_{p(\Gb \given \Zb)} 
		\Big [ h(\Gb) \Big ]
	\right)~
	\prod_{ij}
	\Ncal(z_{ij}; 0, \sigma_z^2) \label{eq:cls-z-prior}
\end{align}
Here, $h$ is the DAG constraint function given in (\ref{eq:trace-exponential}) and $\beta$ is an inverse temperature parameter controlling how strongly the acyclicity is enforced.
As $\beta \rightarrow \infty$, the support of $p_\beta(\Zb)$ reduces to all $\Zb$ that only model valid DAGs.
The Gaussian component ensures that the norm of $\Zb$ is well-behaved.
Traditional graph priors of the form $p(\Gb) \propto f(\Gb)$ that induce, \eg, sparsity of $\Gb$, can be flexibly incorporated into $p_\beta(\Zb)$ by means of an additional factor involving $f(\Gb_\alpha(\Zb))$ or $\EE_{p(\Gb \given \Zb)}[ f(\Gb)]$. 
Unless highlighting a specific point, we omit writing the hyperparameters $\alpha$ and $\beta$ to simplify notation.

\vspacesubcaption
\subsection{Estimators for Gradient-Based Bayesian Inference}\label{sec:dibs-black-box-inference}
\vspacesubcaption
The extended generative model in (\ref{eq:own-generative}) not only allows us to incorporate the notoriously difficult acyclicity constraint into the Bayesian framework.
By rephrasing the posteriors $p(\Gb \given \Dcal)$ and $p(\Gb, \Thetab \given \Dcal)$ in terms of $p(\Zb \given D)$ and $p(\Zb, \Thetab \given D)$, respectively,
it also makes Bayesian structure learning amenable to variational inference techniques that operate in continuous space and rely on the gradient of the unnormalized log posterior, also known as the \emph{score}. The following result provides us with the score of both latent posteriors. Detailed derivations can be found in Appendix \ref{app:proof-lemma-3}.

\begin{proposition}[Latent posterior score]\label{lemma:3}
Under the generative graph model defined in (\ref{eq:own-generative}), the gradient of the log posterior density of $\Zb$ is given by
\begin{align}
	(a) \quad \quad \nabla_{\Zb} \log p(\Zb \given \Dcal)
	&=\nabla_{\Zb} \log p(\Zb) + \frac{\nabla_{\Zb} ~ \EE_{p(\Gb \given \Zb)} \big[ p(\Dcal \given \Gb) \big] }{\EE_{p(\Gb \given \Zb)} \big[ p(\Dcal \given \Gb) \big] }
	\label{eq:cls-target-gradient}
\end{align}
which is relevant when the marginal likelihood $p(\Dcal \given \Gb)$ can be computed efficiently.
In the general case, when inferring the joint posterior of $\Zb$ and $\Thetab$, the gradients of the log posterior are given by
\begin{align}
	(b)& \quad \quad \nabla_{\Zb} \log p(\Zb, \Thetab \given \Dcal)
	=\nabla_{\Zb} \log p(\Zb) + \frac{\nabla_{\Zb} ~ \EE_{p(\Gb \given \Zb)}  \big[p(\Thetab, \Dcal \given \Gb) \big] }{\EE_{p(\Gb \given \Zb)} \big[p(\Thetab, \Dcal \given \Gb) \big]}
	\label{eq:cls-target-gradient-general-z}\\
	(c)& \quad \quad \nabla_{\Thetab} \log p(\Zb, \Thetab \given \Dcal)
	=\frac
		{\EE_{p(\Gb \given \Zb)}  
			\big[ \nabla_{\Thetab} p(\Thetab, \Dcal \given \Gb)   
			\big] }
		{\EE_{p(\Gb \given \Zb)} 
			\big[p(\Thetab, \Dcal \given \Gb) 
			\big]}
	\label{eq:cls-target-gradient-general-theta}
\end{align}
where $p(\Thetab, \Dcal \given \Gb) = p(\Thetab \given \Gb) p(\Dcal \given \Gb, \Thetab)$, which is efficient to compute by construction.
\end{proposition}

\looseness -1 
The expectations have tractable Monte Carlo approximations because
sampling from $p(\Gb \given \Zb)$ in (\ref{eq:graph-latent-model}) 
is simple and parallelizable.
The gradient terms of the form $\nabla_{\Zb} \EE_{p(\Gb \given \Zb)}  [ \,\cdot\,] $, 
which also appear inside $\nabla_{\Zb} \log p(\Zb)$, 
can be estimated using two different techniques, 
depending on the BN model inferred.

\textbf{Differentiable (marginal) likelihood\quad}
Using the Gumbel-softmax trick~\citep{maddison2017concrete,jang2017categorical}, 
we can separate the randomness from $\Zb$ when sampling from $p(\Gb \given \Zb)$ and obtain the following estimator:
\begin{align}
\begin{split}
	 \nabla_{\Zb} \EE_{p(\Gb \given \Zb)}  \big[ p(\Dcal \given \Gb) \big] 
	 &\approx  
	 \EE_{p(\Lb)}  \Big[
	  \nabla_{\Gb} \,  p(\Dcal \given \Gb) \big |_{\Gb = \Gb_{\tau}(\Lb, \Zb)} \cdot
	  \nabla_{\Zb} \, \Gb_{\tau}(\Lb, \Zb)  
	   \Big] \\
\end{split}\label{eq:gumbel-softmax-estimator-g-z}
\end{align} 
where $\Lb \sim \text{Logistic}(0, 1)^{d \times d}$ i.i.d..
The matrix-valued function $\Gb_{\tau}(\cdot)$ is defined elementwise as
\begin{align}
\begin{split}
\Gb_{\tau}(\Lb, \Zb)_{ij}
&:= 
\begin{cases}
		 \sigma_\tau\left(l_{ij} + \alpha \ub_i^\top \vb_j \right) \quad \text{if $i \neq j$}\\
		 0\hspace{81pt}\text{if $i = j$}
\end{cases}
\end{split}
\label{eq:gumbel-softmax-estimator-function}
\end{align} 
\looseness - 1
The estimator applies equally to $p(\Thetab, \Dcal \given \Gb)$ in place of $p(\Dcal \given \Gb)$.
For the estimator to be well-defined, $p(\Dcal \given \Gb)$ or $p(\Thetab, \Dcal \given \Gb)$, respectively, needs to be differentiable with respect to $\Gb$.
More specifically,
the gradient $\nabla_{\Gb} p(\Dcal \given \Gb)$ or $\nabla_{\Gb} p(\Dcal, \Thetab \given \Gb)$ needs to be defined when $\Gb$ lies on the interior of $[0,1]^{d \times d}$ and not at its discrete endpoints.
This depends on the parameterization of the BN model we want to infer.
In case $p(\Dcal \given \Gb)$ or $p(\Thetab, \Dcal \given \Gb)$ is only defined for discrete $\Gb$, it is possible to evaluate $\nabla_{\Gb} p(\Dcal \given \Gb)$ or $\nabla_{\Gb} p(\Thetab, \Dcal \given \Gb)$ using hard Gumbel-max samples of $\Gb$ (\ie, with $\tau = \infty$) and a straight-through gradient estimator. 
Since the DAG constraint $h$ is differentiable, the Gumbel-softmax trick can always be applied inside $\nabla_{\Zb} \log p(\Zb)$.
In practice, we always use $\tau = 1$.

\textbf{Non-differentiable (marginal) likelihood\quad}
In general, $\nabla_{\Gb} p(\Dcal \given \Gb)$ or $\nabla_{\Gb} p(\Dcal, \Thetab \given \Gb)$ depending on the inference task might be not available or ill-defined.
In this setting, the score function estimator provides us with a way to estimate the gradient we need~\citep{williams1992simple}:
\begin{align}
\begin{split}
	 \nabla_{\Zb} \EE_{p(\Gb \given \Zb)}  \big[ p(\Dcal \given \Gb) \big] 
	 &=  \EE_{p(\Gb \given \Zb)}  \Big[ \big ( p(\Dcal \given \Gb)  - b \big ) ~\nabla_{\Zb} \log p(\Gb \given \Zb)  \Big] 
\end{split}\label{eq:sf-estimator-g-z}
\end{align} 
\looseness - 1 The estimator likewise applies for $p(\Thetab, \Dcal \given \Gb)$ in place of $p(\Dcal \given \Gb)$.
Here, $b$ is a constant with respect to $\Gb$ that can be used for variance reduction \citep{mohamed2020monte}, and $\nabla_{\Zb} \log p(\Gb \given \Zb)$ is trivial to compute.
%
The derivations of both (\ref{eq:gumbel-softmax-estimator-g-z}) and (\ref{eq:sf-estimator-g-z}), alongside a more detailed discussion, can be found in Appendix \ref{app:gradient-estimation}.


\vspacecaption
\section{Particle Variational Inference for Structure Learning} \label{sec:dibs-svgd}
\vspacecaptionlow
\looseness -1 In the previous section, we have proposed a {\em differentiable} formulation for {\em Bayesian structure learning (DiBS)} that is agnostic to the form of the local BN conditionals and, more importantly, translates learning discrete graph structures into an inference problem over the continuous variable~$\Zb$.
In the following,
we overcome the remaining challenge of inferring the intractable DiBS posteriors $p(\Zb \given \Dcal)$ and $p(\Zb, \Thetab \given \Dcal)$ by employing Stein variational gradient descent (SVGD) \citep{liu2016stein}, a gradient-based and general purpose variational inference method.
The resulting algorithm infers a particle approximation of the marginal or joint posterior density over BNs given observational data.

\textbf{SVGD for posterior inference\quad}
\looseness - 1
Since Proposition \ref{lemma:3} provides us with the gradient of the latent posterior score functions, we can apply SVGD off-the-shelf.
SVGD minimizes the KL divergence to a target distribution by iteratively \emph{transporting} a set of particles using a sequence of kernel-based transformation steps.
We provide a more detailed overview of SVGD in Appendix \ref{app:background-svgd}.
Following this paradigm for DiBS, we iteratively update a fixed set $\{\Zb^{(m)}\}_{m=1}^M$ or $\{\Zb^{(m)}, \Thetab^{(m)}\}_{m=1}^M$ to approximate $p(\Zb \given \Dcal)$ or $p(\Zb, \Thetab \given \Dcal)$, respectively.
If the BN model we aim to infer has a properly-defined likelihood gradient with respect to $\Gb$, we use the Gumbel-softmax estimator in (\ref{eq:gumbel-softmax-estimator-g-z}) to approximate the posterior score.
Otherwise, we resort to the score function estimator in (\ref{eq:sf-estimator-g-z}).
We use a simple kernel for SVGD: 
\begin{align}
	k\big((\Zb, \Thetab), (\Zb', \Thetab')\big) &:= \exp 
	\left( 
		- \frac{1}{\gamma_z} \lVert \Zb - \Zb' \rVert_2^2
	\right )
	+\exp 
	\left( 
		- \frac{1}{\gamma_\theta} \lVert \Thetab - \Thetab' \rVert_2^2
	\right )
\label{eq:se-kernel}
\end{align}
with bandwidths $\gamma_z, \gamma_\theta$.
For inference of  $p(\Zb \given \Dcal)$, we leave out the second term involving $\Thetab, \Thetab'$.
While $\Zb$ is invariant to orthogonal transformations, more elaborate kernels that are, e.g., invariant to such transforms empirically perform worse in our experiments.

\textbf{Annealing $\alpha$ and $\beta$\quad} 
The latent variable $\Zb$ not only probabilistically models the graph $\Gb$, but can also be viewed as a continuous relaxation of $\Gb$, with $\alpha$ trading off smoothness with accuracy.
As $\alpha$~$\rightarrow$~$\infty$, the sigmoid $\sigma_\alpha(\cdot)$ converges to the unit step function. Hence, as $\alpha$~$\rightarrow$~$\infty$ in the graph model $p_\alpha(\Gb \given \Zb)$ in (\ref{eq:graph-latent-model}), the expectations in Proposition~\ref{lemma:1} simplify to:
\begin{equation}
\begin{split}
	\EE_{p(\Gb \given \Dcal)} \Big [f(\Gb)\Big ] 
	 ~&\rightarrow ~ 
	 \EE_{p(\Zb \given \Dcal)} 
	 \Big [ f\big(\Gb_\infty(\Zb)\big)\Big ]\\
	\displaystyle
	\EE_{p(\Gb, \Thetab \given \Dcal)} \Big [f(\Gb, \Thetab)\Big ] 
	 ~&\rightarrow ~ 
	\EE_{p(\Zb, \Thetab \given \Dcal)} 
	\Big [
		f\big(\Gb_\infty(\Zb), \Thetab\big)
	\Big ]
\end{split}\label{eq:theorem-simplification}
\end{equation}
where $\Gb_\infty(\Zb)$ denotes the single limiting graph implied by $\Zb$~$=$~$[\Ub,\Vb]$ and is defined as
\begin{align}\label{eq:graph-generative-model-matrix-limit}
	\Gb_\infty(\Zb)_{ij} := 
	\begin{cases}
		 1 \quad \text{if $\ub_i^\top \vb_j > 0$ and $i \neq j$}\\
		 0 \quad \text{otherwise}
	\end{cases}
\end{align}
\looseness - 1 
See Appendix \ref{app:proof-lemma-2}. 
In this case, $p_\alpha(\Gb \given \Zb)$ converges to representing only a {\em single graph}.
To be able to invoke this simplification, we anneal $\alpha$~$\rightarrow$~$\infty$ over the iterations of SVGD and, upon termination, convert the latent variables $\Zb$ to the single discrete $\Gb_\infty(\Zb)$.
Furthermore, we similarly let $\beta$~$\rightarrow$~$\infty$ in the latent prior $p_\beta(\Zb)$ over the iterations to enforce that the latent representation of $\Gb$ only models DAGs.
By Equation~(\ref{eq:theorem-simplification}),
the resulting DAGs form a consistent particle approximation of $p(\Gb \given \Dcal)$ or $p(\Gb, \Thetab \given \Dcal)$, respectively.
Algorithm~\ref{alg:DiBS-marginal} summarizes DiBS instantiated with SVGD for inference of $p(\Gb \given \Dcal)$.
The general case of inferring $p(\Gb, \Thetab \given \Dcal)$ is given in Algorithm~\ref{alg:dibs-joint} of  Appendix \ref{app:algo}.
%

\textbf{Single-particle approximation\quad}
\looseness - 1 SVGD reduces to regular gradient ascent for the maximum a posteriori estimate when transporting only a {\em single} particle \citep{liu2016stein}.
In this special case, DiBS with SVGD recovers some of the existing continuous structure learning methods: gradient ascent on a linear Gaussian likelihood solves an optimization problem similar to NOTEARS \citep{zheng2018dags}. The cyclicity penalizer acts analogously.
However, not only does DiBS automatically turn into a full Bayesian approach when using more particles, it is also not limited to settings such as linear Gaussian conditionals, where the adjacency matrix $\Gb$ and the parameters $\Thetab$ can be modeled together by a weighted adjacency matrix.

\textbf{Weighted particle mixture\quad}
In high dimensional settings, it may be beneficial to move beyond a uniform weighting of the inferred particles of BN models to approximate $p(\Gb \given \Dcal)$ or $p(\Gb, \Thetab \given \Dcal)$.
We consider a particle mixture that weights each particle by its unnormalized posterior probability $p(\Gb, \Dcal)$ or $p(\Gb, \Thetab, \Dcal)$, respectively, under the BN model.
While we do not have a strong theoretical justification for the weighting, 
our motivation is that most of the particles in the empirical distribution will be unique as a consequence of the super-exponentially large space of DAGs, which may result in a crude approximation of their posterior probability mass function.
In our experiments, DiBS and its instantiation with SVGD are used interchangeably, and DiBS+ denotes the weighted particle mixture.


\begin{algorithm}[t]
  \footnotesize
  \renewcommand{\algorithmicrequire}{\textbf{Input:}}
  \caption{DiBS with SVGD~\citep{liu2016stein} for inference of $p(\Gb \given \Dcal)$ } \label{alg:DiBS-marginal}
  \hspace*{\algorithmicindent} \textbf{Input:} 
  Initial set of latent particles $\{ \Zb^{(m)}_0 \}_{m=1}^M$, kernel $k$, schedules for $\alpha_t, \beta_t$, and stepsizes $\eta_t$
  \\
  \hspace*{\algorithmicindent} \textbf{Output:} Set of discrete graph particles $\{\Gb^{(m)}\}_{m=1}^M$ approximating $p(\Gb \given \Dcal)$
  \vspace*{0pt}
    \begin{algorithmic}[1] 
    \State Incorporate prior belief of $p(\Gb)$ into $p(\Zb)$ \Comment{See Section \ref{sec:graph-latent-space}}
    \For{iteration $t = 0$ to $T - 1$}
    	\State Estimate score $\nabla_{\Zb} \log p(\Zb \given \Dcal)$ given in (\ref{eq:cls-target-gradient}) for each $\Zb^{(m)}_t$ \Comment{See (\ref{eq:gumbel-softmax-estimator-g-z}) and (\ref{eq:sf-estimator-g-z})}
    	    	
    	\For{particle $m = 1$ to $M$}

        \State $\displaystyle \Zb^{(m)}_{t+1} \leftarrow \Zb^{(m)}_{t} +\eta_t ~ \phib_t(\Zb^{(m)}_t)$ \Comment{SVGD step}
        \Statex \quad \quad \quad \quad   where $\displaystyle \phib_t(\cdot) := \frac{1}{M} \sum_{k=1}^M 
			\Big [  
			k(\Zb^{(k)}_t, \,\cdot\,) ~ \nabla_{\Zb^{(k)}_t} \log p(\Zb^{(k)}_t \given \Dcal )
			+ \nabla_{\Zb^{(k)}_t} k(\Zb^{(k)}_t, \,\cdot\,) 
			\Big ]$

        \EndFor 
    \EndFor 
    \State \Return $\{ \Gb_\infty (\Zb^{(m)}_T) \}_{m=1}^M$ \Comment{See (\ref{eq:theorem-simplification}) and (\ref{eq:graph-generative-model-matrix-limit})}
    \end{algorithmic}
\end{algorithm}
%


\vspacecaption
\section{Evaluation on Synthetic Data} \label{sec:evaluation}
\vspacecaptionlow
\subsection{Experimental Setup\quad} \label{ssec:evaluation-setup}
\textbf{Synthetic data\quad} 
We compare DiBS to a set of related methods in marginal and joint posterior inference of synthetic linear and nonlinear Gaussian BNs.
Our setup follows \citep{zheng2018dags,yu2019daggnn,ng2020role,zheng2020learning}, who consider inferring BNs with Erd{\H{o}}s-R{\'e}nyi and scale-free random structures~\citep{erdos1959random,barabasi1999emergence}.
For each graph, here with $d$~$\in$~$\{20, 50\}$ nodes and $2d$ edges in expectation, we sample a set of ground truth parameters and then generate training, held-out, and interventional data sets.
In all settings, we use $N$~$=$~$100$ observations for inference, emulating the use case of Bayesian structure learning where the uncertainty about the graph structure is significant.

\looseness - 1
\textbf{Graph priors\quad} 
For \erdosrenyi graphs, all methods use the prior $p(\Gb)$~$\propto$~$q^{\lVert \Gb \rVert_1}(1\hspace*{-2pt}-\hspace*{-2pt}q)^{\binom{d}{2} - \lVert \Gb \rVert_1}$, capturing that each edge exists independently w.p.\ $q$~\citep{erdos1959random}. 
\vspace*{-2pt}
For scale-free graphs, we define the prior
$p(\Gb)$~$\propto$~$\prod_{i=1}^d (1 + \lVert \Gb^\top_i\rVert_1 )^{-3}$, analogous to their power law degree distribution $p(\text{deg})$~$\sim$~$ \text{deg}^{-3}$~\citep{barabasi1999emergence}.
Here, $\Gb^\top_i$ is the $i$-th column of the adjacency matrix.
DiBS implements either prior by using the corresponding term above as an additional factor in $p(\Zb)$ with $\Gb := \Gb_\alpha(\Zb)$ (see Section \ref{sec:graph-latent-space}).

\textbf{Metrics\quad}
Since neither density nor samples of the ground truth posteriors are available for BNs of $d \in \{20, 50\}$ variables, we follow the evaluation metrics used by previous work.
We define the \emph{expected structural Hamming distance} to the ground truth graph $\Gb^*$ under the inferred posterior as
\begin{align}
	\EE\text{-SHD}(p, \Gb^*) &:= \sum_\Gb p(\Gb \given \Dcal)  \cdot \text{SHD}(\Gb, \Gb^*)\label{eq:expected-shd}
\end{align}
where $\text{SHD}(\Gb, \Gb^*)$ counts the edge changes that separate the essential graphs representing the MECs of $\Gb$ and $\Gb^*$~\citep{andersson1997characterization,tsamardinos2006max}.
In addition, we follow \citet{friedman2003being} and \citet{ellis2008learning} and compute the \emph{area under the receiver operating characteristic curve (AUROC)} for pairwise edge predictions when varying the confidence threshold under the inferred marginal $p(g_{ij} = 1 \given \Dcal)$.
Finally, following \citet{murphy2001active}, we also evaluate the ability to predict future observations by computing the average \emph{negative (marginal) log likelihood} on 100 held-out observations $\Dcal^{\text{test}}$:
\begin{align}
\begin{split}
	\text{neg.\ LL}(p,  \Dcal^{\text{test}}) &:= - \sum_{\Gb, \Thetab} p(\Gb, \Thetab \given \Dcal)  \cdot \log p(\Dcal^{\text{test}} \given \Gb, \Thetab)
\end{split}\label{eq:heldout-metric}
\end{align}
When inferring $p(\Gb \given \Dcal)$, the corresponding neg.\ MLL metric instead uses $p(\Dcal^{\text{test}} \given \Gb)$.
Analogously, we also compute the \emph{interventional} log likelihoods I-LL and I-MLL, a relevant metric in causal inference~\citep{murphy2001active,cho2016reconstructing}.
Here, an interventional data set $(\Dcal^{\text{int}}, \Ical)$ is instead used to compute $p(\Dcal^{\text{int}} \given \Gb, \Thetab, \Ical)$ and $p(\Dcal^{\text{int}} \given \Gb, \Ical)$ in (\ref{eq:heldout-metric}), respectively.
Scores are the average of 10 interventional data sets.
All reported metrics in this section are aggregated for inference of 30 random synthetic BNs.

\begin{figure}[t]
    \centering
    \vspacefiguretopofpage
    \hspace*{-22pt}\includegraphics{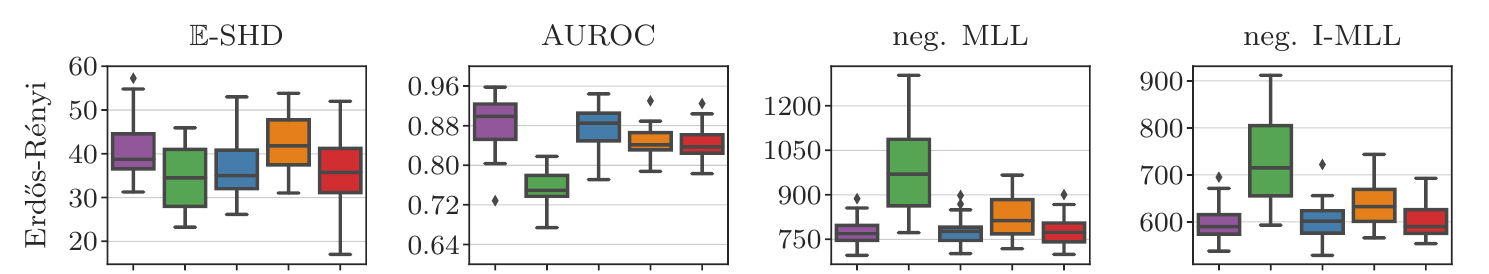}\\
    \vspacefigurebetweenrows
    \hspace*{-22pt}\includegraphics{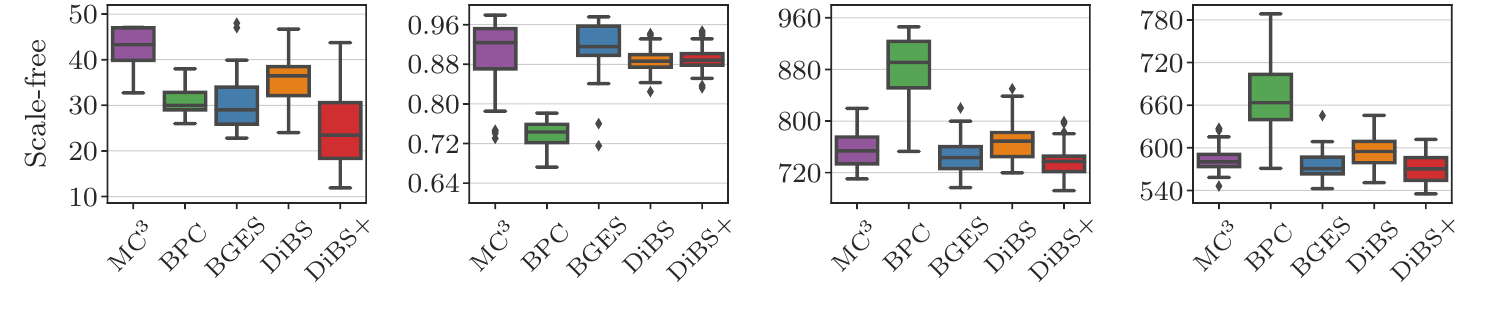} 
    \vspace*{-15pt}
    \vspacefigurecaptiontop \caption{
    \looseness - 1
    Marginal posterior inference of 20-node linear Gaussian BNs using the BGe marginal likelihood. 
    Higher scores on AUROC and lower scores on $\EE\text{-SHD}$, neg.\ MLL, neg.\ I-MLL are preferred. 
    DiBS+ performs competitively across all metrics, in particular $\EE\text{-SHD}$ and neg.\ (I-)MLL.
    \vspacefigurecaption
    }
    \label{fig:bge}
\end{figure}

\looseness - 1 In the remainder, DiBS is always run for 3,000 iterations and with $k$~$=$~$d$ for inference of $d$-variable BNs, which leaves the matrix of edge probabilities unconstrained in rank.
We discard a DiBS particle in the rare case that a returned graph is cyclic. 
Complete details on Gaussian BNs, the evaluation metrics, hyperparameters, and all baselines can be found in Appendix \ref{app:experiments}.

\vspacesubcaption
\subsection{Linear Gaussian Bayesian Networks} \label{ssec:results-lingauss}
\vspacesubcaption

\textbf{Marginal posterior inference\quad}
\looseness - 1 For linear Gaussian BNs, we first evaluate the classical setting of inferring the marginal posterior $p(\Gb \given \Dcal)$ since $p(\Dcal \given \Gb)$ can be computed in closed form.
To this end, we employ the commonly used Bayesian Gaussian Equivalent (BGe) marginal likelihood, which scores Markov equivalent structures equally~\citep{geiger1994learning,geiger2002parameter}. The form of the BGe score requires DiBS to use the score function estimator in (\ref{eq:sf-estimator-g-z}).

We compare DiBS with the nonparametric DAG bootstrap \citep{friedmann1999bootstrap} using the constraint-based PC \citep{spirtes2000causation} and the score-based GES \citep{chickering2003optimal} algorithms (BPC and BGES).
For MCMC, we only consider structure MCMC (\mcmcmc) \citep{giudici2003improving} as a comparison. 
Order MCMC or hybrid DP approaches bound the number of parents and thus often exclude the ground truth graph a priori, especially for scale-free BN structures.
Burn-in and thinning for \mcmcmc are chosen to make the wall time comparable with DiBS run on CPUs. 
In the remainder of the paper, each method uses 30 samples to approximate the posterior over BNs.

\looseness - 1 Figure \ref{fig:bge} summarizes the results for 30 randomly generated BNs with $d$~$=$~$20$ nodes.
We find that DiBS+ performs well compared to the other methods, all of which were specifically developed for the marginal inference scenario evaluated here. DiBS+ appears to be preferable to DiBS.

\textbf{Joint posterior inference\quad}
When inferring the joint posterior $p(\Gb, \Thetab \given \Dcal)$, we can employ a more explicit representation of linear Gaussian BNs, where the conditional distribution parameters are standard Gaussian.
Here, DiBS can leverage the Gumbel-softmax estimator in (\ref{eq:gumbel-softmax-estimator-g-z}) because $p(\Gb, \Thetab \given \Dcal)$ is well-defined when $\Gb$ lies on the interior of $[0,1]^{d \times d}$ (see Appendix \ref{app:gaussian-bns}).
To provide a comparison with DiBS in the absence of an applicable MCMC method, 
we propose two variants of \mcmcmc as baselines. 
Metropolis-Hastings \mcmcmc (M-\mcmcmc) jointly samples parameters and structures, and
Metropolis-within-Gibbs \mcmcmc (G-\mcmcmc) alternates in proposing structure and parameters~\citep{hastings1970monte}.
Moreover, we extend the bootstrap methods by taking the closed-form maximum likelihood estimate \citep{hauser2015jointly} as the posterior parameter sample for a given graph inferred using the BGe score (BPC${}^{*}$ and BGES${}^{*}$), an approach taken in, \eg, causal BN learning \citep{agrawal2019abcdstrategy}.

\begin{figure}[t]
    \centering
    \vspacefiguretopofpage 
    \hspace*{-22pt}\includegraphics{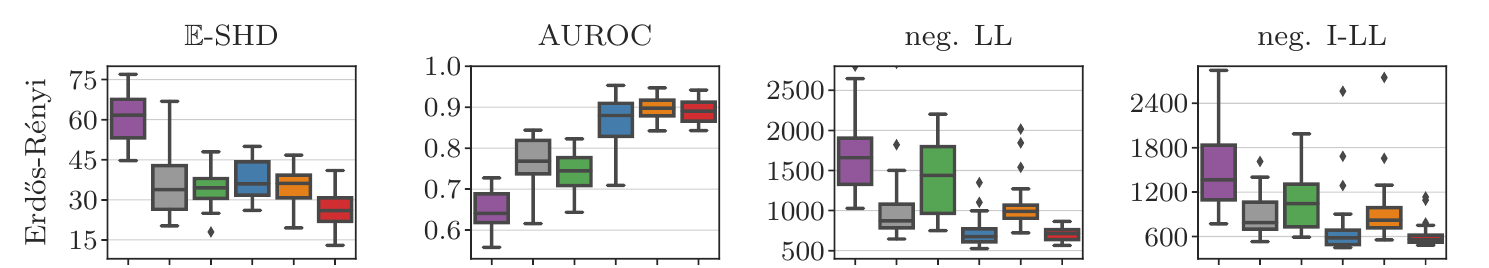}\\
    \vspacefigurebetweenrows
    \hspace*{-22pt}\includegraphics{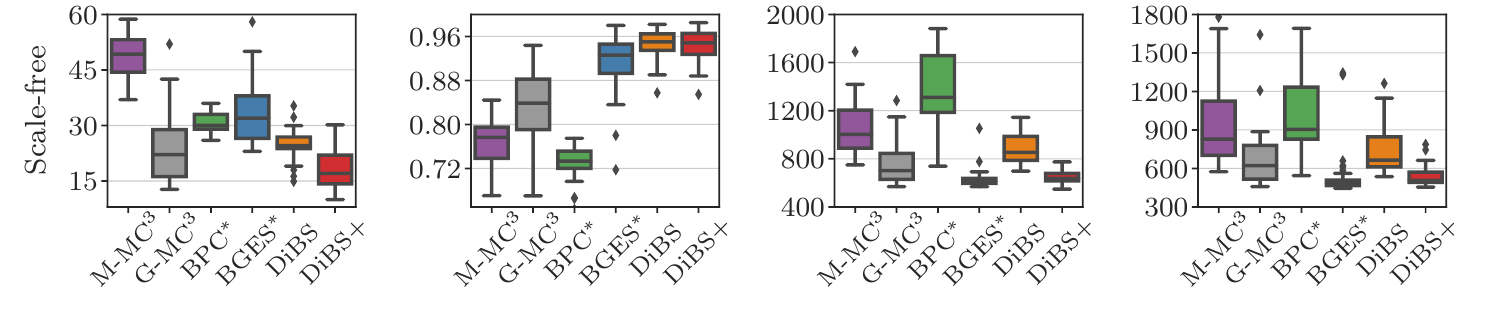} 
    \vspace*{-15pt} \vspacefigurecaptiontop \caption{
    Joint posterior inference of graphs and parameters of linear Gaussian networks with $d$~$=$~$20$ nodes. 
    DiBS+ performs best across all of the metrics.
    BGES${}^{*}$, the next-best alternative, yields substantially worse performance in $\EE$-SHD, \ie, in recovering the overall graph structure and MEC.
    Recall that higher AUROC and lower $\EE\text{-SHD}$, neg.\ LL, and neg.\ I-LL scores are preferable.
    \vspacefigurecaption
    }
    \label{fig:lingauss}
\end{figure}

Figure \ref{fig:lingauss} shows the results for $d$~$=$~$20$ nodes, where $\EE$-SHD and AUROC are computed by empirically marginalizing out the parameters.
We find that DiBS+ is the only considered method that performs well across all of the metrics, often outperforming the baselines by a significant margin.
As for marginal posterior inference of linear Gaussian BNs, DiBS+ performs slightly better than DiBS.

\vspacesubcaption
\subsection{Nonlinear Gaussian Bayesian Networks} \label{ssec:results-nonlinear-gaussian}
\begin{figure}[t]
    \centering
    \vspacefiguretopofpage
    \hspace*{-22pt}\includegraphics{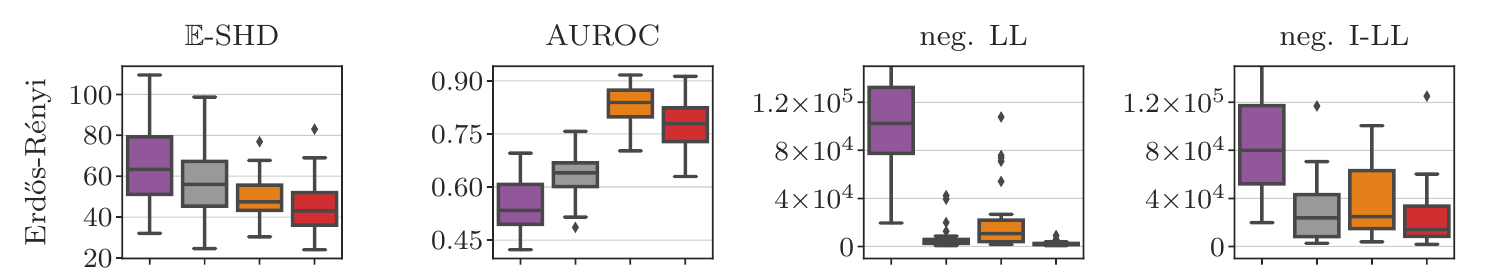}\\
    \vspacefigurebetweenrows \vspace*{-2pt}
    \hspace*{-22pt}\includegraphics{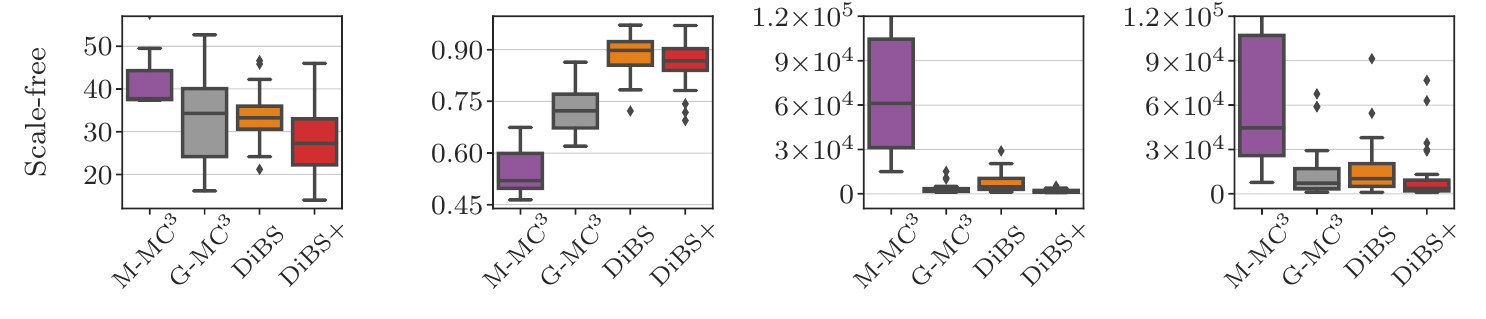} 
    \vspace*{-15pt}  \vspacefigurecaptiontop \vspace*{-5pt} \caption{
    \looseness - 1 Joint posterior inference of nonlinear Gaussian BNs with $d$~$=$~$20$ nodes. Here, the mean of the local conditional distribution of each node is parameterized by a 2-layer neural network with five hidden nodes.
    DiBS and DiBS+ perform favorably across the board, particularly in the graph metrics.
    \vspacefigurecaption  
     \vspace*{-8pt}
    }
    \label{fig:fcgauss}
\end{figure}

\vspacesubcaption
We also consider joint inference of {\em nonlinear} Gaussian BNs where the mean of each local conditional Gaussian is parameterized by a 2-layer dense neural network with five hidden nodes and ReLU activation functions (see Appendix \ref{app:gaussian-bns}).
Since the marginal likelihood does not have a closed form, we are unable to use BPC${}^{*}$ and BGES${}^{*}$ as a means of comparison.
Figure \ref{fig:fcgauss} displays the results for $d$~$=$~$20$ variables, where a given BN model has $|\Thetab|$~$=$~2,220 weights and biases. Analogous to joint inference of linear Gaussian BNs, DiBS and DiBS+ outperform the MCMC baselines across the considered metrics.
To the best of our knowledge, this is the first time that such nonlinear Gaussian BN models have been inferred under the Bayesian paradigm, which opens up exciting avenues in the active learning of more sophisticated causal structures.

Appendix \ref{app:compute} complements Sections \ref{ssec:results-lingauss} and \ref{ssec:results-nonlinear-gaussian} with details on computing time and efficient implementation of DiBS with SVGD, showing that GPU wall times of the above inference experiments lie on the order of seconds or only a few minutes.
In addition, Appendix  \ref{app:additional-results-main} provides results for $d$~$=$~$50$ variables, where DiBS+ likewise performs favorably when jointly inferring $p(\Gb, \Thetab \given \Dcal)$. 
For marginal posterior inference of $p(\Gb \given \Dcal)$ under the BGe marginal likelihood, 
DiBS appears to require more Monte Carlo samples to compensate for the high variance of the score function estimator in this high-dimensional setting.

\vspacesubcaption
\subsection{DiBS with SVGD: Additional Analyses and Ablation Studies}
\vspacesubcaption
Having compared DiBS and its instantiation with SVGD to existing approaches, we finally devote Appendix \ref{app:results-additional} to empirically analyzing some properties of the algorithm.
One of our key results is that, all other things held equal, substituting our inner product model in (\ref{eq:graph-latent-model}) with $p_\alpha(g_{ij} = 1 \given \Zb) = \sigma_{\alpha}(z_{ij})$, where single \emph{scalars} encode the edge probabilities,
results in significantly worse evaluation metrics.

We additionally study the uncertainty quantification in (non)identifiable edge structures and show the effects of reducing the latent dimensionality $k$ or the number of iterations $T$. Our findings suggest that reducing either hyperparameter still allows for competitive posterior approximations and enables trading off posterior inference quality with computational efficiency, \eg, in large-scale applications.

 \vspace*{-2pt}
\section{Application: Inferring Protein Signaling Networks From Cell Data}\label{sec:application}
 \vspace*{-2pt}
\begin{wraptable}{r}{5.7cm}
\vspace*{-13pt}
\small
\caption{
\looseness - 1
Inference of protein signaling pathways with Gaussian BNs.
Metrics are the mean $\pm$ SD of 30 random restarts. 
}\label{tab:results-sachs}
\centering
\vspace*{-4pt}
\hspace*{-2pt}
\begin{threeparttable}
\begin{tabular}{p{0pt} l c c}
	& & $\EE$-\textbf{SHD} & \textbf{AUROC} \\ 
	\cline{1-4}
  	\multirow{5}{*}{$\dagger$} & \mcmcmc             &  34.0 $\pm$ 0.7 & \hspace*{-5pt} 0.616 $\pm$ 0.027    \\
 	& BPC                  & 25.5 $\pm$ 2.3  & \hspace*{-5pt} 0.566 $\pm$ 0.020    \\
 	& BGES                & 33.7 $\pm$ 1.7  & \hspace*{-5pt} 0.641 $\pm$ 0.030    \\
 	& DiBS                 & 37.4 $\pm$ 0.5  & \hspace*{-5pt} \textbf{0.647 $\pm$ 0.047}    \\
 	& DiBS+               & 34.7 $\pm$ 1.5  & \hspace*{-5pt} 0.629 $\pm$ 0.045    \\
 	\cline{1-4}
 	\multirow{4}{*}{$\S$} & M-\mcmcmc\hspace*{-8pt} &  37.3 $\pm$ 3.5  & \hspace*{-5pt}  0.551 $\pm$ 0.078  \\
 	& G-\mcmcmc                &  30.5 $\pm$ 3.2  & \hspace*{-5pt} 0.527 $\pm$ 0.067  \\
 	& DiBS                      &  23.4 $\pm$ 0.5  & \hspace*{-5pt} 0.598 $\pm$ 0.052 \\
 	& DiBS+ \hspace*{-5pt}      &  22.9 $\pm$ 2.7  & \hspace*{-5pt} 0.557 $\pm$ 0.052 \\
 	\cline{1-4}
 	\multirow{4}{*}{$\P$} & M-\mcmcmc\hspace*{-8pt} & 25.2 $\pm$ 3.0  & \hspace*{-5pt} 0.526 $\pm$ 0.084 \\
 	& G-\mcmcmc                & 35.1 $\pm$ 3.2  & \hspace*{-5pt} 0.540 $\pm$ 0.080  \\
 	& DiBS                      & \textbf{22.6 $\pm$ 0.5}  & \hspace*{-5pt} 0.577 $\pm$ 0.039  \\
 	& DiBS+ \hspace*{-5pt}     &  22.8 $\pm$ 1.9  & \hspace*{-5pt} 0.535 $\pm$ 0.041 \\ 
 	\cline{1-4}
\end{tabular}
\begin{tablenotes}
\scriptsize
\vspace*{1pt}
\item[\hspace*{-6pt}$\dagger$] \hspace*{-9pt} Linear Gaussian BN; graph only via BGe marginal lik.\
\item[\hspace*{-6pt}$\S$] \hspace*{-9pt} Linear Gaussian BN; graph and parameters jointly
\item[\hspace*{-6pt}$\P$] \hspace*{-9pt} Nonlinear Gaussian BN; graph and parameters jointly
\end{tablenotes}
\end{threeparttable}
\vspace*{-20pt}
\end{wraptable}

A widely used benchmark in structure learning is the proteomics data set by \citet{sachs2005causal}.
The data contain $N$~$=$~$7,466$ continuous measurements of $d$~$=$~$11$ proteins involved in human immune system cells as well as an established causal network of their signaling interactions. 

We infer both linear and nonlinear Gaussian BNs with \erdosrenyi graph priors exactly as 
in Section \ref{sec:evaluation}.
The AUROC results in Table~\ref{tab:results-sachs} indicate that the posterior by DiBS under the BGe model provides the most calibrated edge confidence scores. 
Not penalizing model complexity as much, the marginal BGe posterior of DiBS (DiBS+) averages a high expected number of 39.0 (35.0) edges, compared to 12.7 (14.2) and 12.6 (14.2) for its joint posteriors over linear and nonlinear BNs, respectively. Appendix~\ref{app:application-details} provides further details and analyses on this matter.
The $\EE$-SHD scores show that, among all the methods, DiBS is closest in structure to the consensus network when performing joint inference with nonlinear Gaussian BNs. 
This further highlights the need for nonlinear conditionals and joint inference of the graph and parameters in complex real-world settings.

\vspacecaption 
\section{Conclusion}\label{sec:discussion}
\vspacecaptionlow 

\looseness - 1
We have presented a general, fully differentiable approach to inference of posterior distributions over BNs. Our framework is based on a continuous latent representation of DAGs, whose posterior can be equivalently inferred---without loss of generality and using existing black box inference methods.
While we have used SVGD~\citep{liu2016stein} for this purpose, our general approach could also be instantiated with, \eg, gradient-based sampling methods that rely on the score of the target density~\citep{neal2011mcmc,welling2011bayesian}.
This may improve upon the asymptotic runtime of DiBS with SVGD, which scales quadratically in the number of sampled particles.
We expect that our end-to-end approach can be extended to handle missing and interventional data as well as to amortized contexts, where rich unstructured data is available.

\textbf{Broader impact\quad}
\looseness -1 
Our work is relevant to any scientific discipline that aims at inferring the (causal) structure of a system or reasoning about the effects of interventions.
If algorithms and decisions are grounded in the structural understanding of a data-generating system and take into account the epistemic uncertainty, we expect them to be more robust and have fewer unforeseen side-effects. 
However, the assumptions allowing for a causal interpretation of DAGs, \eg, the absence of unmeasured confounders, are often untestable and to be taken with care~\citep{dawid2010beware}, particularly in safety-critical and societally-sensitive applications. 
Hence, while potential misuse can never be ruled out, our presented method predominantly promises positive societal and scientific impact.

\begin{ack} 
This  project received  funding  from  the Swiss National Science Foundation under NCCR Automation under grant agreement 51NF40 180545, the  European Research  Council  (ERC)  under  the European  Union’s Horizon  2020  research  and  innovation  program  grant agreement no.\ 815943, and was supported with compute resources by Oracle Cloud Services. 
This work was also supported by the German Federal Ministry of Education and Research (BMBF): T{\"u}bingen AI Center, FKZ: 01IS18039B, and by the Machine Learning Cluster of Excellence, EXC number 2064/1 – Project number 390727645.
We  thank  Nicolo Ruggeri and Guillaume Wang for their valuable feedback.
\end{ack}

{
\small 
\bibliographystyle{unsrtnat} 
\bibliography{references.bib}
}

\numberwithin{equation}{section}
\appendix

\newpage 

\section{Proofs of the Main Results}\label{app:proofs}

\subsection{Proposition \ref{lemma:1}} \label{app:proof-lemma-1}

\paragraph{Proof} For ease of understanding, we recall that the generative model in (\ref{eq:own-generative}) factorizes the joint distribution as 
$p(\Zb, \Gb, \Thetab, \Dcal) = p(\Zb) p(\Gb \given \Zb) p(\Thetab \given \Gb)  p(\Dcal \given \Gb, \Thetab)$.
First, let us consider case (a), the setting where the marginal likelihood $p(\Dcal \given \Gb)$ can be computed in closed form.
We get
\begin{align}
	\EE_{p(\Gb \given \Dcal)} f(\Gb)
	&= \sum_{\Gb}  p(\Gb \given \Dcal) f(\Gb) \\
	&= \sum_{\Gb} \frac{p(\Gb, \Dcal) f(\Gb)}{p(\Dcal)}  \\
	&= \sum_{\Gb} \int_{\Zb} \frac{p(\Zb, \Gb, \Dcal) f(\Gb)}{p(\Dcal)} d\Zb   \\
	&= \sum_{\Gb} \int_{\Zb} \frac{p(\Zb)p(\Gb \given \Zb)  p(\Dcal \given \Gb)  f(\Gb)}{p(\Dcal)} d\Zb \\
	&\quad\quad\quad \text{by the generative model in (\ref{eq:own-generative})}  \nonumber \\
	\nonumber \\
	&= \sum_{\Gb} \int_{\Zb} \frac{p(\Zb \given \Dcal) p(\Gb \given \Zb) p(\Dcal \given \Gb)  f(\Gb)}{p(\Dcal \given \Zb)} d\Zb \\
	&\quad\quad\quad \text{since}~ 
	p(\Zb \given \Dcal) = \frac{p(\Zb)p(\Dcal \given \Zb)}{p(\Dcal)}  \nonumber \\
	\nonumber \\
	&= \int_{\Zb} p(\Zb \given \Dcal) ~ \frac{\sum_{\Gb}  p(\Gb \given \Zb) p(\Dcal \given \Gb)  f(\Gb)}{p(\Dcal \given \Zb)} d\Zb \quad \quad \text{rearranging} \\
	\nonumber \\
	&= \int_{\Zb} p(\Zb \given \Dcal) ~ \frac{\sum_{\Gb}  p(\Gb \given \Zb) p(\Dcal \given \Gb)  f(\Gb)}{\sum_{\Gb} p(\Gb, \Dcal \given \Zb)} d\Zb  \\
	&\quad\quad\quad \text{by the law of total probability}\nonumber \\
	\nonumber \\
	&= \int_{\Zb} p(\Zb \given \Dcal) ~ \frac{\sum_{\Gb}  p(\Gb \given \Zb) p(\Dcal \given \Gb)  f(\Gb)}{\sum_{\Gb} p(\Gb \given \Zb) p(\Dcal \given \Gb) } d\Zb  \\
	&\quad\quad\quad \text{expanding $p(\Gb, \Dcal \given \Zb)$ by the generative model in (\ref{eq:own-generative})}\nonumber \\	
	\nonumber \\
	&= \EE_{p(\Zb \given \Dcal)} \Bigg [  
	\frac
	{\EE_{p(\Gb \given \Zb)} 
	\big [
		 f(\Gb) p(\Dcal \given \Gb)  
	\big ]
	}
	{\EE_{p(\Gb \given \Zb)} 
	\big [
		 p(\Dcal \given \Gb)  
	\big ]
	}
	\Bigg ]
\end{align}

as desired for (a).
	
Finally, let us consider (b), the general case. The derivation essentially follows the same ideas as for (a) but does not marginalize out $\Thetab$.
\begin{align}
	&\EE_{p(\Gb, \Thetab \given D)} f(\Gb, \Thetab)\\
	&= \sum_{\Gb} \int_{\Thetab} p(\Gb, \Thetab \given \Dcal) f(\Gb, \Thetab) d\Thetab  \\
	&= \sum_{\Gb} \int_{\Thetab} \frac{p(\Gb, \Thetab, \Dcal) f(\Gb, \Thetab)}{p(\Dcal)} d\Thetab  \\
	&= \sum_{\Gb} \int_{\Thetab} \int_{\Zb} \frac{p(\Zb, \Gb, \Thetab, \Dcal) f(\Gb, \Thetab)}{p(\Dcal)} d\Zb d\Thetab   \\
	&= \sum_{\Gb} \int_{\Thetab} \int_{\Zb} \frac{p(\Zb)p(\Gb \given \Zb) p(\Thetab \given \Gb) p(\Dcal \given \Gb, \Thetab)  f(\Gb, \Thetab)}{p(\Dcal)} d\Zb d\Thetab \\
	&\quad\quad\quad \text{by the generative model in (\ref{eq:own-generative})}  \nonumber \\
	\nonumber \\
	&= \sum_{\Gb} \int_{\Thetab} \int_{\Zb} \frac{p(\Zb, \Thetab \given \Dcal) p(\Gb \given \Zb) p(\Thetab \given \Gb) p(\Dcal \given \Gb, \Thetab)  f(\Gb, \Thetab)}{p(\Dcal, \Thetab \given \Zb)} d\Zb d\Thetab \\
	&\quad\quad\quad \text{since}~ 
	p(\Zb, \Thetab \given \Dcal) = \frac{p(\Zb)p(\Thetab, \Dcal \given \Zb)}{p(\Dcal)}  \nonumber \\
	\nonumber \\
	&= \int_{\Thetab} \int_{\Zb} p(\Zb, \Thetab \given \Dcal) ~ \frac{\sum_{\Gb}  p(\Gb \given \Zb) p(\Thetab \given \Gb) p(\Dcal \given \Gb, \Thetab)  f(\Gb, \Thetab)}{p(\Thetab, \Dcal \given \Zb)} d\Zb d\Thetab  \\
	&\quad\quad\quad \text{rearranging} \nonumber \\
	\nonumber \\
	&= \int_{\Thetab} \int_{\Zb} p(\Zb, \Thetab \given \Dcal) ~ \frac{\sum_{\Gb}  p(\Gb \given \Zb) p(\Thetab \given \Gb) p(\Dcal \given \Gb, \Thetab)  f(\Gb, \Thetab)}{\sum_{\Gb} p(\Gb, \Thetab, \Dcal \given \Zb)} d\Zb d\Thetab  \\
	&\quad\quad\quad \text{by the law of total probability}\nonumber \\
	\nonumber \\
	&= \int_{\Thetab} \int_{\Zb} p(\Zb, \Thetab \given \Dcal) ~ \frac{\sum_{\Gb}  p(\Gb \given \Zb) p(\Thetab \given \Gb) p(\Dcal \given \Gb, \Thetab)  f(\Gb, \Thetab)}{\sum_{\Gb} p(\Gb \given \Zb) p(\Thetab \given \Gb) p(\Dcal \given \Gb, \Thetab) } d\Zb d\Thetab  \\
	&\quad\quad\quad \text{expanding $p(\Gb, \Thetab, \Dcal \given \Zb)$ by the generative model in (\ref{eq:own-generative})}\nonumber \\	
	\nonumber \\
	&= \EE_{p(\Zb, \Thetab \given \Dcal)} \Bigg [  
	\frac
	{\EE_{p(\Gb \given \Zb)} 
	\big [
		 f(\Gb, \Thetab) p(\Thetab \given \Gb) p(\Dcal \given \Gb, \Thetab)  
	\big ]
	}
	{\EE_{p(\Gb \given \Zb)} 
	\big [
		 p(\Thetab \given \Gb) p(\Dcal \given \Gb, \Thetab)
	\big ]
	}
	\Bigg ]
\end{align}

which is the statement in (b). $\qed$

\subsection{Proposition \ref{lemma:3}} \label{app:proof-lemma-3}

\paragraph{Proof} We will derive the gradients of the unnormalized posterior since
\begin{align}
	\nabla_{\Zb} \log p(\Zb \given \Dcal) = \nabla_{\Zb} \log p(\Zb, \Dcal) -  \nabla_{\Zb} \log p(\Dcal) =  \nabla_{\Zb} \log p(\Zb, \Dcal)
\end{align}
and analogously for the other two expressions.
Through straightforward manipulation and using the identity 
$\nabla_{\xb} \log f(\xb) 	= \nabla_{\xb} f(\xb) / f(\xb)$,
we obtain
\begin{align}
	\nabla_{\Zb} \log p(\Zb, \Dcal)
	&=\nabla_{\Zb} \log p(\Zb) + \nabla_{\Zb}  \log  p(\Dcal \given \Zb)\\
	&=\nabla_{\Zb} \log p(\Zb) + \frac{\nabla_{\Zb} ~  p(\Dcal \given \Zb)}{p(\Dcal \given \Zb)} \\
	&=\nabla_{\Zb} \log p(\Zb) + \frac{\nabla_{\Zb} \big [ \sum_{\Gb} p(\Gb \given \Zb) p(\Dcal \given \Gb) \big] }{\sum_{\Gb} p(\Gb \given \Zb) p(\Dcal \given \Gb) } \\
	&=\nabla_{\Zb} \log p(\Zb) + \frac{\nabla_{\Zb} ~ \EE_{p(\Gb \given \Zb)} [ p(\Dcal \given \Gb) ] }{\EE_{p(\Gb \given \Zb)} [ p(\Dcal \given \Gb) ] }
\end{align}
Analogously, we get
\begin{align}
	\nabla_{\Zb} \log p(\Zb, \Thetab, \Dcal)
	&=\nabla_{\Zb} \log p(\Zb) + \nabla_{\Zb}  \log  p(\Thetab, \Dcal \given \Zb)\\
	&=\nabla_{\Zb} \log p(\Zb) + \frac{\nabla_{\Zb} ~ p(\Thetab, \Dcal \given \Zb)}{p(\Thetab, \Dcal \given \Zb)} \\
	&=\nabla_{\Zb} \log p(\Zb) + \frac{\nabla_{\Zb} \big [ \sum_{\Gb}  p(\Gb \given \Zb) p(\Thetab, \Dcal \given \Gb) \big ] }{\sum_{\Gb} p(\Gb \given \Zb) p(\Thetab, \Dcal \given \Gb) } \\
	&=\nabla_{\Zb} \log p(\Zb) + \frac{\nabla_{\Zb} ~ \EE_{p(\Gb \given \Zb)}  [p(\Thetab, \Dcal \given \Gb) ] }{\EE_{p(\Gb \given \Zb)} [p(\Thetab, \Dcal \given \Gb) ]}
\end{align}
Lastly, using the same ideas as above, we arrive at
\begin{align}
	\nabla_{\Thetab} \log p(\Zb, \Thetab, \Dcal)
	&=\nabla_{\Thetab} \log p(\Zb) + \nabla_{\Thetab}  \log  p(\Thetab, \Dcal \given \Zb)\\
	&=\frac{\nabla_{\Thetab} ~ p(\Thetab, \Dcal \given \Zb)}{p(\Thetab, \Dcal \given \Zb)} \\
	&=\frac{\nabla_{\Thetab} \big [ \sum_{\Gb} p(\Gb \given \Zb) p(\Thetab, \Dcal \given \Gb) \big ]}{\sum_{\Gb} p(\Gb \given \Zb) p(\Thetab, \Dcal \given \Gb) } \\
	&=\frac{\sum_{\Gb} p(\Gb \given \Zb) \nabla_{\Thetab} p(\Thetab, \Dcal \given \Gb)}{\sum_{\Gb} p(\Gb \given \Zb) p(\Thetab, \Dcal \given \Gb) } \\
	&=\frac
		{\EE_{p(\Gb \given \Zb)}  
			\big[\nabla_{\Thetab} p(\Thetab, \Dcal \given \Gb)
			\big] }
		{\EE_{p(\Gb \given \Zb)} 
			[p(\Thetab, \Dcal \given \Gb)]} \qed
\end{align}

In the above, without any additional factor modeling a prior belief over graphs, the score of the latent prior $p_\beta(\Zb)$ as defined in (\ref{eq:cls-z-prior}) is given by
\begin{align}
    \nabla_{\Zb} \log p_\beta(\Zb) 
    &= - \beta ~ \nabla_{\Zb}~ \EE_{p(\Gb \given \Zb)} [ h(\Gb)  ] - \frac{1}{\sigma_z^2} \Zb
\end{align}

\paragraph{Practical considerations}
Estimating expectations of the form $\EE_{p(\Gb \given \Zb)}[f(\Gb)]$ with Monte Carlo sampling can be numerically challenging when $f$ are probability densities and thus often close to zero.
In practice, we recommend the log-sum-exp trick for applying Proposition \ref{lemma:3}. Let us define
\begin{align}\label{eq:logsumexp}
    \logsumexp_{m=1}^M\{x^{(m)}\} := \log \left ( \sum_{m=1}^M \exp\big(x^{(m)}\big) \right )
\end{align}
For $M$ Monte Carlo samples $\Gb^{(m)} \sim p(\Gb \given \Zb)$, we can rewrite the estimator for the expectation as
\begin{align}
    \EE_{p(\Gb \given \Zb)} [ f(\Gb)] 
    &\approx \frac{1}{M} \sum_{m=1}^M f(\Gb^{(m)})
    = \exp \left ( \log \left ( \sum_{m=1}^M f\big(\Gb^{(m)}\big) \right ) - \log M \right )\\
    &= \exp \left ( \logsumexp_{m=1}^M\Big\{ \log f \big(\Gb^{(m)}\big) \Big\} - \log M \right) 
\end{align}
\looseness - 1 
Computing $\logsumexp$ can be made numerically stable by subtracting and adding $\max_m \{ x^{(m)} \}$ before and after applying $\logsumexp$ to $\{x^{(m)}\}$, respectively.
Stable $\logsumexp$ can be extended to handle negative-valued $f$ inside the expectation, \eg, for the gradient of $f$, and to the ratios of expectations in Proposition \ref{lemma:3}.

\subsection{Equations (\ref{eq:theorem-simplification}) and (\ref{eq:graph-generative-model-matrix-limit})}  \label{app:proof-lemma-2}

\paragraph{Proof} The sigmoid function converges to the unit step function, \ie $\sigma_\alpha(x) \rightarrow \mathbbm{1}[x > 0]$ as $\alpha \rightarrow \infty$.
Hence, the edge probabilities $\Gb_\alpha(\Zb)$ defined in \eq{eq:graph-generative-model-matrix} converge to a (binary) matrix $\Gb$ as $\alpha \rightarrow \infty$.
Extending the notation of \eq{eq:graph-generative-model-matrix}, we will denote this single limiting graph implied by $\Zb$ as $\Gb_\infty(\Zb)$ where
\begin{align*}
	\Gb_\infty(\Zb)_{ij} := 
	\begin{cases}
		 1 \quad \text{if $\ub_i^\top \vb_j > 0$ and $i \neq j$}\\
		 0 \quad \text{otherwise}
	\end{cases}
\end{align*}
The above implies that when the temperature parameter $\alpha$ is annealed to $\infty$, the probability mass function and correspondingly the expectation simplify:
\begin{align}
\begin{split}
	\text{As}~\alpha \rightarrow \infty: ~~
	&p_\alpha(\Gb \given \Zb) \rightarrow \mathbbm{1} [ \Gb = \Gb_\infty(\Zb) ]\\
	&\EE_{p_\alpha(\Gb \given \Zb)}[f(\Gb)] \rightarrow f(\Gb_\infty(\Zb))\label{eq:g-given-w-convergence}
\end{split}
\end{align}
Again, let us first consider case (a).
Starting with Proposition \ref{lemma:1}(a) in the first step, we can use the above insight to simplify the inner expectations:
\begin{align}
	\EE_{p(\Gb \given \Dcal)} 
	\Big [f(\Gb)\Big ]  
	&= 
	\EE_{p(\Zb \given \Dcal)} \Bigg [  
	\frac
	{\EE_{p_\alpha(\Gb \given \Zb)} 
	\big [
		 f(\Gb) p(\Dcal \given \Gb)  
	\big ]
	}
	{\EE_{p_\alpha(\Gb \given \Zb)} 
	\big [
		 p(\Dcal \given \Gb)  
	\big ]
	}
	\Bigg ]\\
	&\hspace*{-17pt}\stackrel{\alpha \rightarrow \infty}{\longrightarrow} ~~\EE_{p(\Zb \given \Dcal)} 
	\Bigg [
		 \frac{f(\Gb_{\infty}(\Zb))p(\Dcal \given \Gb_{\infty}(\Zb))}{p(\Dcal \given \Gb_{\infty}(\Zb))}
	\Bigg ] \\
	&= \EE_{p(\Zb \given \Dcal)} 
	\Big [
		f(\Gb_{\infty}(\Zb))
	\Big ]
\end{align}
Analogously, we get for the general case (b):
\begin{align}
	\hspace*{-7pt}\EE_{p(\Gb, \Thetab \given \Dcal)}  \Big [f(\Gb, \Thetab)\Big ] 
	&=
	\EE_{p(\Zb, \Thetab \given \Dcal)} \Bigg [  
	\frac
		{\EE_{p_\alpha(\Gb \given \Zb)}~ \big [ f(\Gb, \Thetab)p(\Thetab \given \Gb)p(\Dcal \given \Gb, \Thetab) \big ] }
		{\EE_{p_\alpha(\Gb \given \Zb)}~ \big [ p(\Thetab \given \Gb)p(\Dcal \given \Gb, \Thetab) \big ] }
	\Bigg ]
	\\
	&\hspace*{-17pt}\stackrel{\alpha \rightarrow \infty}{\longrightarrow} ~~\EE_{p(\Zb, \Thetab \given \Dcal)} 
	\Bigg [
		 \frac{f(\Gb_{\infty}(\Zb), \Thetab)p(\Thetab \given \Gb_{\infty}(\Zb)) p(\Dcal \given \Gb_{\infty}(\Zb), \Thetab)}{p(\Thetab \given \Gb_{\infty}(\Zb)) p(\Dcal \given \Gb_{\infty}(\Zb), \Thetab)}
	\Bigg ] \\
	&= \EE_{p(\Zb, \Thetab \given \Dcal)} 
	\Big [
		f(\Gb_{\infty}(\Zb), \Thetab)
	\Big ] \qed
\end{align}

\section{Gradient Estimation for Bayesian Inference}\label{app:gradient-estimation}

To derive the expressions for the gradient estimators in (\ref{eq:gumbel-softmax-estimator-g-z}) and (\ref{eq:sf-estimator-g-z}) for both the marginal likelihood and the likelihood cases, we will use a generic function $f(\Gb)$ as a placeholder for either $p(\Dcal \given \Gb)$ or $p(\Dcal \given \Gb, \Thetab)$, since the results hold for general densities $f(\Gb)$.

\subsection{Gumbel-Softmax Estimator for the Likelihood Gradient} \label{app:proof-gumbel-grad}
In general, for a Bernoulli random variable $X$ with $p(X = 1) = q$, it holds that 
\begin{align}
	X \stackrel{d}{=} \mathbbm{1}\big[G_1 + \log q > G_0 + \log(1 - q)\big]\label{eq:gumbel-softmax-bernoulli}
\end{align}
when $G_0, G_1 \sim \text{Gumbel}(0, 1)$.
This is the Gumbel-{\em max} trick.
Since the unit step function $\mathbbm{1}[\cdot]$ does not have an informative gradient, \citet{maddison2017concrete} and \citet{jang2017categorical} have proposed to use the sigmoid function $\sigma_\tau(\cdot)$ with parameter $\tau$ as a soft relaxation of $\mathbbm{1}[\cdot]$.

Using this so-called Gumbel-{\em softmax} trick, we can reparameterize the entries of $\Gb$ under the graph model in (\ref{eq:graph-latent-model}).
Starting from the Gumbel-max equality in (\ref{eq:gumbel-softmax-bernoulli}), we rearrange the inequality inside the indicator  into the form ``$> 0$'' and apply the sigmoid relaxation with parameter $\tau$. We obtain the following soft relaxation for each entry of $\Gb$:
\begin{align}
	g_{ij} 
	&\approx \sigma_\tau\Big(G_1 - G_0 + \log \sigma_\alpha(\ub_i^\top \vb_j) - \log(1 - \sigma_\alpha(\ub_i^\top \vb_j))\Big)\\
	&= \sigma_\tau\left(L + \log \left ( \frac{\sigma_\alpha(\ub_i^\top \vb_j)}{\sigma_\alpha(- \ub_i^\top \vb_j)} \right)\right)\\
	&= \sigma_\tau\left(L + \log \left  ( \frac{\exp(\alpha \ub_i^\top \vb_j)}{\exp(\alpha \ub_i^\top \vb_j) + 1} \frac{\exp(\alpha \ub_i^\top \vb_j) + 1}{1} \right)\right)\\
	&= \sigma_\tau\left(L + \log \left  ( \exp(\alpha \ub_i^\top \vb_j) \right)\right)\\
	&= \sigma_\tau\left(L + \alpha \ub_i^\top \vb_j \right)
\end{align}
where $L \sim \text{Logistic}(0, 1)$ since $L \stackrel{d}{=} G_1 - G_0$ when $G_0, G_1 \sim \text{Gumbel}(0,1)$.
For $i=j$, we set $g_{ij} := 0$ by default in accordance with the graph model in (\ref{eq:graph-latent-model}).
Since this allows us to separate the randomness in sampling from the distribution $p(\Gb \given \Zb)$ from the values of $\Zb$, we can move the gradient operator inside the expectation and obtain the estimator given in (\ref{eq:gumbel-softmax-estimator-g-z}):
\begin{align}
\begin{split}
	 \nabla_{\Zb} \EE_{p(\Gb \given \Zb)}  \big[ f(\Gb) \big] 
	 &\approx  
	 \EE_{p(\Lb)}  \Big[
	  \nabla_{\Zb} \, f(\Gb_{\tau}(\Lb, \Zb))   
	   \Big] \\
	 &=  \EE_{p(\Lb)}  \Big[
	  \nabla_{\Gb} \, f(\Gb) \big |_{\Gb = \Gb_{\tau}(\Lb, \Zb)} \cdot
	  \nabla_{\Zb} \, \Gb_{\tau}(\Lb, \Zb)  
	   \Big] \\
\end{split}
\end{align} 

While the reparameterization trick generally provides a lower variance estimate of the gradient, the form in (\ref{eq:gumbel-softmax-estimator-g-z}) is biased when $\tau < \infty$ because we use a soft relaxation of the true distribution.
In addition, the estimator in (\ref{eq:gumbel-softmax-estimator-g-z}) requires that $\nabla_{\Gb} p(\Dcal \given \Gb)$ or $\nabla_{\Gb} p(\Thetab, \Dcal \given \Gb)$ is available, depending on the inference task.
In case  $p(\Dcal \given \Gb)$ or $p(\Thetab, \Dcal \given \Gb)$ is only defined for discrete $\Gb$, it is possible to evaluate $\nabla_{\Gb} p(\Dcal \given \Gb)$ or $\nabla_{\Gb} p(\Thetab, \Dcal \given \Gb)$ using hard Gumbel-max samples of $\Gb$ (\ie, with $\tau = \infty$). 
As before, however, one would use soft Gumbel-softmax samples in $ \nabla_{\Zb} \, \Gb_{\tau}(\Lb, \Zb)$ to obtain an informative gradient.
Lastly, we can use this estimator to approximate the score of the latent prior $\nabla_{\Zb}\log p(\Zb)$ given in (\ref{eq:cls-z-prior}) because the acyclicity constraint $h(\Gb)$ is differentiable with respect to~$\Gb$.
In practice, the log-sum-exp trick described in Section \ref{app:proof-lemma-3} as well as the score function identity 
$\nabla_{\xb} f(\xb) = f(\xb) \nabla_{\xb} \log f(\xb) $ should be used for numerically stable computation of the estimator.

\subsection{Score Function Estimator for the Likelihood Gradient} \label{app:proof-score-function-grad}
To arrive at the estimator in (\ref{eq:sf-estimator-g-z}), we expand the  gradient expression as
\begin{align}
	 \nabla_{\Zb} \EE_{p(\Gb \given \Zb)}  \big[ f(\Gb) \big] \label{eq:proof-sf-estimator-definition}
	 &=  \sum_\Gb f(\Gb) \nabla_{\Zb}  p(\Gb \given \Zb) 
	 =  \sum_\Gb f(\Gb) p(\Gb \given \Zb) \nabla_{\Zb}  \log p(\Gb \given \Zb)  \\
	 &=  \EE_{p(\Gb \given \Zb)}  \Big[  f(\Gb) ~\nabla_{\Zb} \log p(\Gb \given \Zb)  \Big] 
\end{align} 
where (\ref{eq:proof-sf-estimator-definition}) uses the fact that $\nabla_{\Zb}  \log p(\Gb \given \Zb)   = \nabla_{\Zb}  p(\Gb \given \Zb)  /  p(\Gb \given \Zb)$.
Finally, we recall the well-known property of the score function that
$\EE_{p(\Gb \given \Zb)}  
	 [ \nabla_{\Zb} \log p(\Gb \given \Zb) ] 
= \mathbf{0}$.
Due to this, for any constant $b$ as written in (\ref{eq:sf-estimator-g-z}), the estimator is unbiased because the additional term involving $b$ has zero expectation.
The constant can be used to reduce the variance of the Monte Carlo estimator \citep{mohamed2020monte}. In our experiments, we always use $b = 0$.

\section{Background: Stein Variational Gradient Descent}\label{app:background-svgd}
This section describes Stein variational gradient descent (SVGD) by \citet{liu2016stein}.
The overview is meant as supplementary material for Section \ref{sec:dibs-svgd}, where we propose to use SVGD for inferring the DiBS posteriors $p(\Zb \given \Dcal)$ and $p(\Zb, \Thetab \given \Dcal)$.
In contrast to sampling-based MCMC or optimization-based variational inference methods, SVGD iteratively {\em transports} a fixed set of particles to closely match a target distribution, akin to the gradient descent algorithm in optimization.
We refer the reader to \citet{liu2016stein} for additional details.

Let $p(\xb)$ with $\xb$~$\in$~$\Xcal$ be a differentiable density that we want to sample from, \eg, to estimate an expectation.
Starting from a smooth reference density $q(\xb)$, SVGD aims to find a one-to-one transform $\tb: \Xcal \mapsto \Xcal$ such that the transformed density $q_{[\tb]}(\widetilde{\xb})$ with $\widetilde{\xb} := \tb(\xb)$ minimizes the KL-divergence to $p$.
In particular, \citet{liu2016stein} propose to use the incremental transform 
\begin{align}\label{eq:svgd-background-transform}
    \tb(\xb) = \xb + \eta ~\phib(\xb) 
\end{align}
When $|\eta|$ is sufficiently small, the Jacobian of $\tb$ has full rank and $\tb$ is one-to-one.
The key result by \citet{liu2016stein} links the incremental transform $\tb$ in (\ref{eq:svgd-background-transform}) to prior work on reproducing kernel Hilbert spaces (RKHSs). The authors show that if $\phib$ lies in the unit ball of the RKHS induced by a kernel $k$, then the transform $\tb$ maximizing the descent on the KL divergence from $q_{[\tb]}$ to $p$ uses an incremental update $\phib$ proportional to
\begin{align}\label{eq:svgd-background-optimal-step}
    \phib_{q,p}^*(\cdot) = \EE_{q(\xb)} \left [
        k(\xb, \cdot) \nabla_\xb \log p(\xb)^\top + \nabla_\xb k(\xb, \cdot)
    \right ] 
\end{align}
This suggests an iterative procedure of repeatedly applying the update of (\ref{eq:svgd-background-transform})
\vspace*{-2pt} 
with $\phib = \phib_{q,p}^*(\cdot)$ from (\ref{eq:svgd-background-optimal-step}) to a finite set of randomly initialized particles $\{ \xb^{(m)} \}_{m=1}^M$. 
At each iteration $t$, the $m$-th particle $\xb^{(m)}$ is then deterministically updated according to: 
\vspace*{-3pt} 
\begin{align}\label{eq:svgd-background-step}
\begin{split}
    &\xb_{t+1}^{(m)} \leftarrow \xb_{t}^{(m)}  + \eta_t ~\phib(\xb_{t}^{(m)}) \\
    &\text{where}~~ \phib(\xb) =  \frac{1}{M} \sum_{k=1}^M 
			\Big [  
			k(\xb_{t}^{(k)}, \xb) ~ \nabla_{\xb_{t}^{(k)}} \log p(\xb_{t}^{(k)})
			+ \nabla_{\xb_{t}^{(k)}} k(\xb_{t}^{(k)}, \xb) 
			\Big ] 
\end{split}
\end{align}
For sufficiently small step sizes $\eta_t$, the sequence of particles eventually converges, in which case the transform $\tb$ reduces to the identity mapping. The particle update in (\ref{eq:svgd-background-step}) consists of a gradient ascent term driving the particles to high-density regions, and a term involving $\nabla_{\xb} k(\xb, \cdot)$ that acts as a repulsive force between particles, preventing them from collapsing into the modes of $p(\xb)$.

\section{General Algorithm}\label{app:algo}

\begin{algorithm}[H]
  \footnotesize
  \renewcommand{\algorithmicrequire}{\textbf{Input:}}
  \caption{DiBS with SVGD~\citep{liu2016stein} for inference of $p(\Gb, \Thetab \given \Dcal)$}
  \label{alg:dibs-joint}
 \hspace*{\algorithmicindent} \textbf{Input:} 
  Initial latent and parameter particles $\{ (\Zb^{(m)}_0, \Thetab^{(m)}_0) \}_{m=1}^M$,  kernel $k$, schedules for $\eta_t$, $\alpha_t, \beta_t$
  \\
  \hspace*{\algorithmicindent} \textbf{Output:} Set of graph and parameter particles $\{(\Gb^{(m)}, \Thetab^{(m)})\}_{m=1}^M$ approximating $p(\Gb, \Thetab \given \Dcal)$
  \vspace*{0pt}
    \begin{algorithmic}[1] 
    \State Incorporate prior belief of $p(\Gb)$ into $p(\Zb)$ \Comment{See Section \ref{sec:graph-latent-space}}
    \For{iteration $t = 0$ to $T - 1$}
    	\State Estimate score $\nabla_{\Zb} \log p(\Zb, \Thetab \given \Dcal)$ given in (\ref{eq:cls-target-gradient-general-z}) for each $\Zb^{(m)}_t$ \Comment{See  (\ref{eq:gumbel-softmax-estimator-g-z}) and  (\ref{eq:sf-estimator-g-z})}
    	\vspace*{-2pt}
    	\State Estimate score $\nabla_{\Thetab} \log p(\Zb, \Thetab \given \Dcal)$ given in (\ref{eq:cls-target-gradient-general-theta}) for each $\Thetab^{(m)}_t$ 
    	    	
    	\For{particle $m = 1$ to $M$}

        \State $\displaystyle \Zb^{(m)}_{t+1} \leftarrow \Zb^{(m)}_{t} +\eta_t ~ \phib^{\Zb}_t\big(\Zb^{(m)}_t, \Thetab^{(m)}_t \big)$ \Comment{SVGD step}
        \Statex \hspace*{30pt} where $\displaystyle \phib^{\Zb}_t(\cdot, \cdot) := \frac{1}{M} \sum_{k=1}^M 
			\Big [  
			k\big((\Zb^{(k)}_t, \Thetab^{(k)}_t), \,(\cdot, \cdot)\,\big) ~ \nabla_{\Zb^{(k)}_t} \log p\big(\Zb^{(k)}_t, \Thetab^{(k)}_t \given \Dcal \big)$
		\Statex \hspace*{220pt} $
			+ \nabla_{\Zb^{(k)}_t} k\big((\Zb^{(k)}_t, \Thetab^{(k)}_t), \,(\cdot, \cdot)\,\big) 
			\Big ]$
		\vspace*{-3pt}
        \State $\displaystyle \Thetab^{(m)}_{t+1} \leftarrow \Thetab^{(m)}_{t} +\eta_t ~ \phib^{\Thetab}_t\big(\Zb^{(m)}_t, \Thetab^{(m)}_t \big)$ 
         \Statex \hspace*{30pt} where $\phib^{\Thetab}_t(\cdot, \cdot)$ is analogous to $\phib^{\Zb}_t(\cdot, \cdot)$ but using gradients $\nabla_{\Thetab_t^{(k)}}$ instead of $\nabla_{\Zb_t^{(k)}}$ 
        \EndFor 
    \EndFor 
    \State \Return $\{ ( \Gb_\infty (\Zb^{(m)}_T), \Thetab^{(m)}_T ) \}_{m=1}^M$ \Comment{See (\ref{eq:theorem-simplification}) and (\ref{eq:graph-generative-model-matrix-limit})}
    \end{algorithmic}
\end{algorithm}

\section{Experimental Details}\label{app:experiments}

\subsection{Gaussian Bayesian Networks}\label{app:gaussian-bns}
In our experiments, we consider Bayesian networks with Gaussian local conditional distributions of each variable given its parents.
For both linear or nonlinear Gaussian BNs, which will be defined presently, the generative model for synthetic data simulation as well as the parameter prior used for joint inference are set to standard Gaussian distributions. We fix the observation noise to $\sigma^2 = 0.1$ for all nodes both during synthetic data generation and joint posterior inference, rendering the causal structure fully identifiable \citep{peters2014identifiability}.

\paragraph{Linear}
Analogous to linear regression, linear Gaussian BNs model the mean of a given variable as a linear function of its parents:
\begin{align}
\begin{split}
	p(\xb \given \Gb, \Thetab) 
	&= 	\prod_{i=1}^d \Ncal (x_i ; \thetab_i^\top \xb_{\pa(i)}, \sigma^2)
	\\
	\hspace*{-20pt}\text{or}\quad p(\xb \given \Gb, \Thetab) 
	&= \Ncal \big(\xb ; (\Gb \circ \Thetab )^\top \xb, \sigma^2 \Ib \big)
\end{split} \label{eq:lingauss}
\end{align}
where ``$\circ$'' denotes elementwise multiplication.
In our experiments, DiBS uses the second parameterization in (\ref{eq:lingauss}) to allow for a constant dimensionality of the conditional distribution parameters~$\Thetab$ and make the likelihood well-defined for the Gumbel-softmax estimator in (\ref{eq:gumbel-softmax-estimator-g-z}).

When inferring the marginal posterior $p(\Gb \given \Dcal)$ for linear Gaussian BNs, we follow the predominant choice in the literature and employ the \emph{Bayesian Gaussian Equivalent (BGe)} marginal likelihood, under which Markov equivalent structures are scored equally~\citep{geiger1994learning,geiger2002parameter}.
Details on the computation of the BGe score are provided by \citet{kuipers2014addendum}.
Following the notation of \citet{geiger2002parameter} and \citet{kuipers2014addendum}, we use the standard effective sample size hyperparameters $\alpha_\mu = 1$ and $\alpha_\omega = d + 2$ as well as the diagonal form of the Wishart inverse scale matrix for the Normal-Wishart parameter prior underlying the BGe score.
\paragraph{Nonlinear}
The interaction between variables $\xb$ can straightforwardly be extended to be \emph{non\-linear}, \eg, using neural networks.
In Section \ref{ssec:results-nonlinear-gaussian}, 
we follow \citet{zheng2020learning} and consider (fully connected) feed-forward neural networks (FFNs) of the form
\begin{align}\label{eq:dense-nn}
\begin{split}
	\text{FFN}( \, \cdot \, ; \Thetab) & : \RR^{d} \rightarrow \RR \\
	\text{FFN}(\ub; \Thetab) & := 
	\Thetab^{(L)}
		\sigma\Big(
		\dots 
			\Thetab^{(2)}
			\sigma\big(\Thetab^{(1)}
				\ub
				+ \thetab_b^{(1)}
			\big)
			+ \thetab_b^{(2)}
		\dots
	\Big)
	+ \thetab_b^{(L)}
\end{split}
\end{align}
with weights $\Thetab^{(l)} \in \RR^{d_{l} \times d_{l - 1}}$, biases $\thetab^{(l)}_b \in \RR^{d_{l}}$, and elementwise activation function $\sigma: \RR \rightarrow \RR$.
\citet{zheng2020learning} show that
the class of fully connected neural networks in (\ref{eq:dense-nn}) that do {\em not} depend on the value of $u_k$ is equivalent to the class of fully connected neural networks in (\ref{eq:dense-nn}) where the $k$-th column of $\Thetab^{(1)}$ equals zero.
This insight allows us to define a nonlinear Gaussian BN parameterized by a fully connected neural network:
\begin{align}
\begin{split}
	p(\xb \given \Gb, \Thetab) 
	&= 	\prod_{i=1}^d \Ncal \Big(x_i ; ~\text{FFN}\big(\Gb^\top_i \circ \xb; \Thetab_i \big), ~\sigma^2 \Big)\\
\end{split} \label{eq:nn-nonlingauss}
\end{align}
As required for a BN, each variable is independent of its non-descendants given its parents. 
The mask representation in (\ref{eq:lingauss}) and (\ref{eq:nn-nonlingauss}) is equivalent to the concept of a structural gate used by \citet{kalainathan2018structural}.
Note that the conditional distribution parameters for a single nonlinear Gaussian BN of the form in (\ref{eq:nn-nonlingauss}) contain the weights and biases of $d$ different neural networks, one for the local conditional distribution of each node.

\subsection{Evaluation metrics}\label{app:evaluation-metrics}
We provide additional details on the evaluation metrics used throughout the paper. Bayesian structure learning beyond five variables is notoriously difficult to evaluate since the ground truth posterior is not accessible. We hence rely and build on the metrics established in the literature.

\paragraph{Expected structural Hamming distance}
The structural Hamming distance $\text{SHD}(\Gb, \Gb^*)$ between two graphs $\Gb$ and $\Gb^*$  counts the edge changes that separate the essential graphs representing the MECs of $\Gb$ and $\Gb^*$~\citep{andersson1997characterization,tsamardinos2006max}.
We define the expected structural Hamming distance to the ground truth graph $\Gb^*$ under the inferred posterior as
\begin{align*}
	\EE\text{-SHD}(p, \Gb^*) &:= \sum_\Gb p(\Gb \given \Dcal)  \cdot \text{SHD}(\Gb, \Gb^*)  \; .
\end{align*}
Empirically, the $\EE$-SHD is similar to the  $L_1$ edge error used by \citet{tong2001active} and \citet{murphy2001active}, but also takes into account the MEC.
The $\EE$-SHD is computed via Monte Carlo estimation of the expectation using samples from the posterior. Note that the DAG bootstrap variants and DiBS+ use the weighted mixture rather than the empirical distribution of samples.
In the joint inference setting, we empirically marginalize out $\Thetab$ to obtain $p(\Gb \given \Dcal)$.

\paragraph{Receiver operating characteristic}
The marginal posterior $p(\Gb \given \Dcal)$ provides a confidence estimate $p(g_{ij} = 1 \given \Dcal)$ for whether a given edge $(i,j)$ is present in the ground truth DAG $\Gb^*$.
Recall that the marginal posterior edge probability $p(g_{ij} = 1 \given \Dcal)$ is the posterior mean of an indicator for the presence of that edge, \ie, $p(g_{ij} = 1 \given \Dcal) = \EE_{p(\Gb \given \Dcal)} \mathbbm{1}[g_{ij} = 1]$, which amounts to counting the proportion of graphs with $g_{ij} = 1$ (and to weighted counting for the DAG bootstrap variants and DiBS+). 
The receiver operating characteristic (ROC) curve is then obtained by viewing the presence of each of the $d^2$ possible edges in a $d$-node graph as a binary classification task and varying the decision threshold from 0 to 1 under our confidence estimates $p(g_{ij} = 1 \given \Dcal)$.
The area under the receiver operating characteristic curve (AUROC) evaluates faithful uncertainty quantification of the posterior. In general, random guessing achieves an AUROC of 0.5 in expectation; a perfect classifier achieves an AUROC of 1.

\paragraph{Held-out log likelihood} 
We also evaluate the ability to predict future observations by computing the average negative log likelihood on 100 held-out observations $\Dcal^{\text{test}}$ defined as
\begin{align*}
	\text{neg.\ LL}(p,  \Dcal^{\text{test}}) &:= - \sum_{\Gb, \Thetab} p(\Gb, \Thetab \given \Dcal)  \cdot \log p(\Dcal^{\text{test}} \given \Gb, \Thetab) \; .
\end{align*}
As for $\EE$-SHD, the neg.\ LL is a posterior mean and thus computed via Monte Carlo estimation using samples from the posterior.
When inferring $p(\Gb \given \Dcal)$, the corresponding neg.\ MLL metric uses the marginal likelihood $p(\Dcal^{\text{test}} \given \Gb)$ instead of the likelihood $p(\Dcal^{\text{test}} \given \Gb, \Thetab)$.

\paragraph{Held-out log interventional likelihood}
Lastly, to capture relevant performance metrics in causal inference~\citep{murphy2001active,cho2016reconstructing}, we  also compute the negative \emph{interventional} log likelihood.
Given an interventional data set $(\Dcal^{\text{int}}, \Ical)$, the interventional likelihood is given by 
\begin{align}
\begin{split}
	 p(\Dcal^{\text{int}} \given \Gb, \Thetab, \Ical) = \prod_{\xb^{(n)} \in \Dcal} ~ \prod_{\substack{j=1 \\ j \notin \Ical}}
	 p(x^{(n)}_j \given \xb^{(n)}_{\Gb_j}, \thetab_j)  
\end{split} \label{eq:definition-interventional-likelihood}
\end{align}
where $\xb_{\Gb_j}$ are the values of the parents of variable $j$ in $\Gb$, and $\thetab_j$ parameterizes the local conditional distribution of $j$.
The neg.\ I-LL and neg.\ I-MLL metrics are defined analogous to the neg.\ LL and neg.\ MLL in (\ref{eq:heldout-metric}) but use the interventional likelihood in (\ref{eq:definition-interventional-likelihood}) instead of the observational likelihood.
For marginal posterior inference, we likewise use the interventional marginal likelihood $p(\Dcal^{\text{int}} \given \Gb, \Ical)$ instead.
In our experiments, we obtain interventional data $(\Dcal^{\text{int}}, \Ical)$ by randomly selecting 10\% of the variables and clamping them to zero in the ground-truth data-generating process. The reported neg.\ I-LL and I-MLL scores are the average of 10 different interventional data sets with $|\Dcal^\text{int}| = 100$.

\subsection{Hyperparameters}\label{app:hparams}
In all evaluations, DiBS is run for 3,000 iterations and uses the simple linear constraint schedule $\beta_t$~$:=$~$t$. 
At $t$~$=$~$0$, the initial latent particles $\{\Zb_0\}_{m=1}^M$ and parameter particles $\{\Thetab_0\}_{m=1}^M$ are initialized by sampling from their prior distributions. 
For the step size schedule $\eta_t$, we use the adaptive learning rate method RMSProp with learning rate 0.005.
We always use $128$ samples for Monte Carlo estimation of the gradients.
Finally, the bandwidths $\gamma_z, \gamma_\theta$ of the kernel in (\ref{eq:se-kernel}) and the slope of a linear schedule $\alpha_t$ are chosen in separate held-out instances of each setting in Section \ref{sec:evaluation} and are listed in Table \ref{tab:dibs-params}.
As illustrated by the application in Section \ref{sec:application}, the provided hyperparameters can be expected to apply to problem settings of comparable magnitude.

While the latent variable scale $\sigma_z$ can in principle be set arbitrarily, we always set $\sigma_z = 1/\sqrt{k}$ in the prior $p(\Zb)$, which makes the norm in the SE kernel given in (\ref{eq:se-kernel}) roughly invariant with respect to the latent dimension $k$, ignoring the acyclicity term.
This follows from the fact that $||\ub||^2 \sim \text{Gamma}(k/2, 2\sigma_z^2)$ when $u_i \sim \Ncal(0,\sigma_z^2)$, in which case $\EE[||\ub||^2] = k\sigma_z^2$.

\begin{table}[ht]
\caption{
DiBS hyperparameter choices for $\alpha_t$ and bandwidths $\gamma_z, \gamma_\theta$. Here, $\widetilde{\alpha}$ denotes the slope in the linear schedule $\alpha_t := \widetilde{\alpha} t $.
}\label{tab:dibs-params}
\centering
\vspace{10pt}
\begin{tabular}{l c c c c}
	Model & $d$ & $\widetilde{\alpha}$ & $\gamma_z$ & $\gamma_\theta$ \\ \hline
 	\multirow{2}{*}{BGe}    & 20 & 2  & 2 & -\\
	& 50 & 2 & 50 & -\\ \hline
	\multirow{2}{*}{Linear Gaussian}   & 20 & 0.2  & 5 & 500\\
	& 50 & 0.02  & 15 & 1,000 \\ \hline
	\multirow{2}{*}{Nonlinear Gaussian}    & 20 & 0.02 & 5 & 1,000 \\
    & 50 & 0.01 & 15 & 2,000 \\ \hline
\end{tabular}
\centering
\end{table}

\subsection{Baseline Methods} \label{app:baseline-details}

\paragraph{Structure MCMC (\mcmcmc, M-\mcmcmc, G-\mcmcmc)}
Designed for inference of the marginal posterior $p(\Gb \given \Dcal)$,
structure MCMC \citep{madigan1995bayesian,giudici2003improving} performs sampling in the space of DAGs by adding, deleting, and reversing one edge at a time without violating acyclicity.
The acceptance probability of a proposed graph $\Gb'$ is given by %
\begin{align}\label{eq:mcmcmc-acceptance}
    \min \left \{ 1, 
    \frac
    {|\Ncal(\Gb\hphantom{'})|\cdot p(\Dcal \given \Gb') p(\Gb')}
    {|\Ncal(\Gb')|\cdot p(\Dcal \given \Gb\hphantom{'}) p(\Gb\hphantom{'})}
    \right \}
\end{align}
where $\Gb$ is the current particle and $\Ncal(\Gb)$ is the collection of DAGs reachable from $\Gb$ with one edge change.
Following \citep{giudici2003improving}, the ratio of neighborhoods is approximated to equal one, which allows for only computing $\Ncal(\Gb')$ when accepting $\Gb'$.
We implement \mcmcmc using the efficient ancestor matrix trick for finding acyclic proposals \citep{giudici2003improving}.
For marginal inference under the BGe marginal likelihood, we compute the Bayes factor in (\ref{eq:mcmcmc-acceptance}) by only taking into account the affected node families.

For all of \mcmcmc, M-\mcmcmc, and G-\mcmcmc, we specify a burn-in period of 100k samples and then collect a sample every 10k steps, which makes the wall time of \mcmcmc and DiBS on CPUs comparable.
Both M-\mcmcmc and G-\mcmcmc use a simple Gaussian random walk proposal for the parameters, respectively, with scale selected to roughly obtain an acceptance rate of 0.2 in each setting~\citep{sherlock2010random}, when feasible in combination with the graph proposal.

\paragraph{Nonparametric DAG bootstrap (BPC, BGES, BPC${}^{*}$, BGES${}^{*}$)}
The nonparametric DAG bootstrap \citep{friedmann1999bootstrap} performs model averaging by bootstrapping the observations $\Dcal$ to yield a collection of synthetic data sets, each of which is used to learn a single graph, here using the GES and PC algorithms \citep{chickering2003optimal,spirtes2000causation}.
The collection of unique single graphs approximates the posterior by weighting each graph by its unnormalized posterior probability in (\ref{eq:bayesian-structure-learning-marg-posterior}), analogous to DiBS+.
The closed-form maximum likelihood parameter estimate for linear Gaussian BNs with known $\Gb$, which is used by BPC${}^{*}$ and BGES${}^{*}$ to allow approximating the joint posterior,
is provided by \citet{hauser2015jointly}.
For joint posterior inference, BPC${}^{*}$ and BGES${}^{*}$ use $p(\Gb, \Thetab, \Dcal)$ rather than $p(\Gb, \Dcal)$ for weighting the inferred BN models.

Since the GES and PC algorithms only return essential graphs, \ie, MECs, we favor them in computing the AUROC score. We orient a predicted undirected edge correctly when a ground truth edge exists and only count a falsely predicted undirected edge as a single mistake.
The held-out likelihood metrics given in (\ref{eq:heldout-metric}) are computed for a random consistent DAG extension of the essential graph \citep{dor1992simple}.
Enumerating the possibly exponential number of DAGs in an MEC is infeasible in general \citep{he2015counting}.
Implementations of the PC and GES algorithms are given by the \texttt{CausalDiscoveryToolbox} \citep{kalainathan2019causal}, which is published under an MIT Licence and executes their commonly used R implementations.


\section{Efficient Implementation and Computing Resources}\label{app:compute}
DiBS and SVGD operate on continuous tensors and solely rely on Monte Carlo estimation and parallel gradient ascent-like updates.
Thus, our inference framework allows for a highly efficient implementation using vectorized operations, automatic differentiation, just-in-time compilation, and hardware acceleration.
For this purpose, we implement DiBS with \texttt{JAX}~\citep{jax2018github}, which is published under an Apache Licence.
Our code is publicly available at:~
\href{https://github.com/larslorch/dibs}{\color{blue}\texttt{https://github.com/larslorch/dibs}}.

Table \ref{tab:compute} summarizes the computing time of DiBS on GPU and CPUs for a superset of the inference tasks on BNs with \erdosrenyi structures in Section \ref{sec:evaluation}.
The GPU wall times for medium-sized inference tasks of up to around 20 nodes and 30 particles lie on the order of seconds or a few minutes.
None of the baseline methods considered in this work are comparable in terms of efficiency and usage of modern hardware accelerators.
Moreover, in larger inference problems than evaluated here, \texttt{JAX} would directly allow for the computations of DiBS to be performed in a distributed fashion, \eg, by updating batches of SVGD particles across multiple GPU devices.
Note that BGe wall times are relatively slow because the closed-form marginal likelihood involves determinants~\citep{geiger1994learning,geiger2002parameter}.

\begin{table}[H]
\caption{\looseness - 1 Wall times of DiBS for the hyperparameters described in Section \ref{app:hparams}.
Times are the mean of 10 random restarts. 
$M$ denotes the number of particles, $d$ the number of nodes, and GPU/CPU the processing backend of \texttt{JAX}.
We used one \textsc{NVIDIA GeForce RTX 2080 Ti} GPU or one full 2.70GHz \textsc{Intel XeonGold 6150} CPU node to measure GPU and CPU wall time, respectively, for each run.
Experiments marked by a dash exceeded the GPU memory.
The main experiments of Sections \ref{sec:evaluation} and~\ref{sec:application} were performed in bulk on Oracle \textsc{BM.Standard.E2.64} CPU machines and no GPUs. 
}
    \label{tab:compute}
    \vspace{15pt}
    \centering
    
    \begin{adjustbox}{max width=\linewidth}
    \begin{tabular}{lcccccccc}
    \multirow{4}{*}{\textbf{Model}} & &
    \multicolumn{7}{c}{\textbf{Wall time (min)}} \\
    \cmidrule{3-9}
    & & \multicolumn{3}{c}{\textbf{GPU}} & &  \multicolumn{3}{c}{\textbf{CPU}}\\
    \cmidrule{3-5}\cmidrule{7-9}
    & $d=$ & {10} & {20} & {50} & & {10} & {20} & {50}\\
    \cmidrule{1-9}
    \multicolumn{9}{l}{\textbf{BGe}} \\
    \cmidrule{1-9}
    $M = 10$  && 0.349  & 0.892  & 8.111 && 3.337  & 18.251  & 216.844 \\
    $M = 30$ && 0.609 & 2.084  & ---  && 9.380 & 52.203  & 659.106  \\
    \cmidrule{1-9} 
    \multicolumn{9}{l}{\textbf{Linear Gaussian}} \\
    \cmidrule{1-9}    
    $M = 10$  && 0.370 & 0.603 & 1.571  && 1.410  & 3.738 & 21.531\\
    $M = 30$ && 0.612 & 1.153 & 3.975  && 4.703 & 13.163 & 74.914 \\
    \cmidrule{1-9}
    \multicolumn{9}{l}{\textbf{Nonlinear Gaussian}} \\
    \cmidrule{1-9}
    $M = 10$ && 2.702 & 6.139 & 24.117 && 10.019 & 27.749 & 128.820  \\
    $M = 30$ && 7.476 & 17.667 & --- && 28.130 & 79.628 & 388.992  \\
    \cmidrule{1-9}
    \end{tabular}
    \end{adjustbox}
\end{table}

\newpage 

\section{Results for Gaussian Bayesian networks with $d$~$=$~$50$ variables} \label{app:additional-results-main}
\vspace*{-10pt}

\begin{figure}[!ht]
    \centering
    \hspace*{-23pt}
    \includegraphics{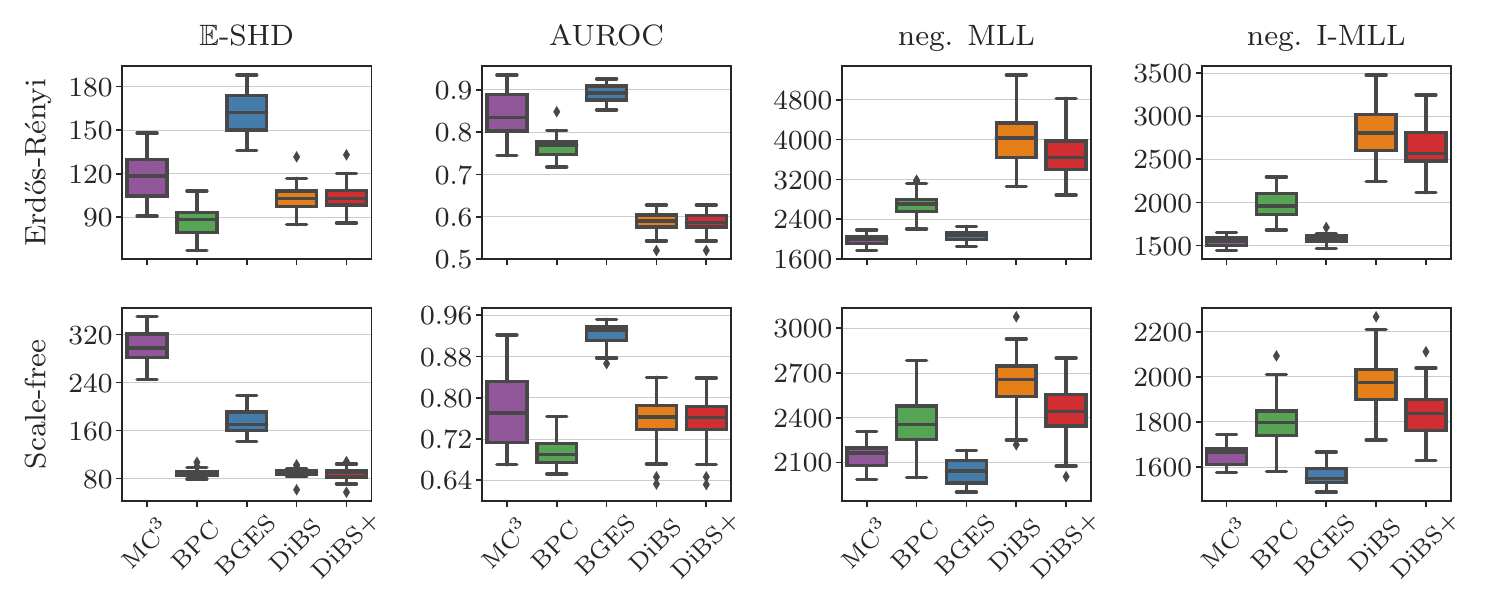}
    \vspace*{-15pt}
    \vspacefigurecaptiontop
    \caption{
    Marginal posterior inference of linear Gaussian BNs with $d$~$=$~$50$ variables using the BGe marginal likelihood. 
    The metrics are aggregated for 30 random BNs of each graph type. 
    While DiBS and DiBS+ are competitive in the structural $\EE$-SHD and AUROC metrics, we find that the baselines specifically designed for marginal posterior inference perform favorably in the likelihood-based metrics.
    We hypothesize that this is due to the high variance incurred by the score function estimator that DiBS needs to use in the marginal inference setting under the BGe model (cf.\ Section \ref{ssec:results-lingauss}).
    To reach comparable results with DiBS in this high-dimensional setting, the DiBS score function gradient estimator may require more than the default 128 Monte Carlo samples used here. 
    \vspacefigurecaption 
    }
    \label{fig:bge50}
\end{figure}

\begin{figure}[!ht]
    \centering
    \hspace*{-22pt}
    \includegraphics{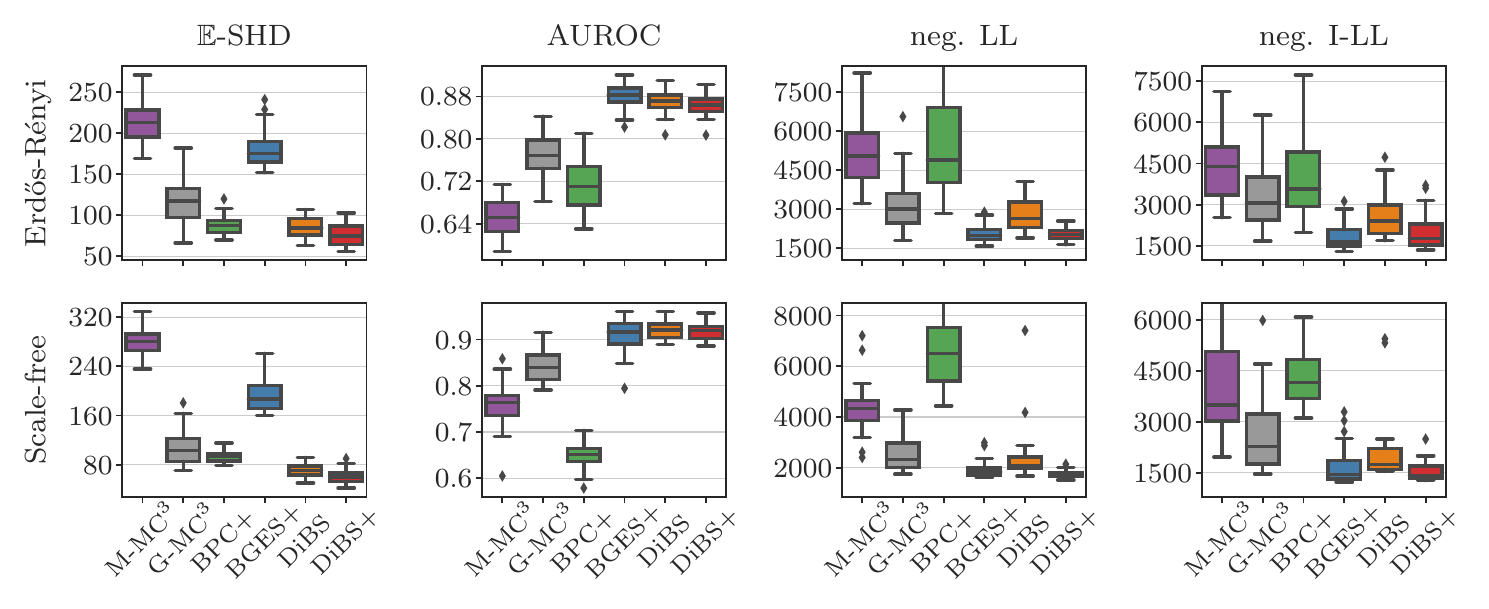}
    \vspace*{-15pt}
    \vspacefigurecaptiontop
    \caption{
    Joint posterior inference of linear Gaussian BNs with $d$~$=$~$50$ variables.
    The first and second rows show the aggregate metrics for inference of 30 random BNs with \erdosrenyi and scale-free structures, respectively.
    Analogous to inference for $d$~$=$~$20$ variables, DiBS+  outperforms all alternatives to joint posterior inference of the graph and the conditional distribution parameters across the metrics.
    \vspacefigurecaption 
    }
    \label{fig:lingauss50}
\end{figure}

\newpage

\begin{figure}[!ht]
    \centering
    \vspace*{-10pt}
    \hspace*{-23pt}
    \includegraphics{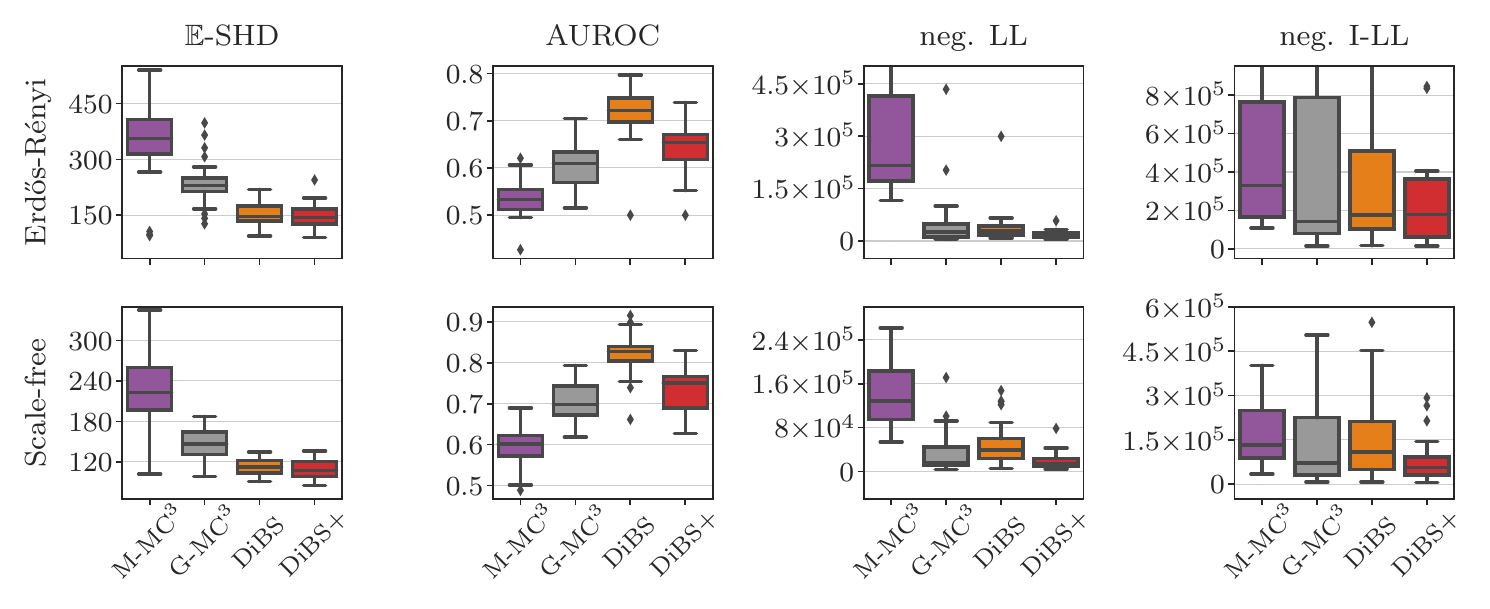}
    \vspace*{-15pt}
    \vspacefigurecaptiontop
    \caption{
    \looseness - 1 Joint posterior inference of nonlinear Gaussian BNs with $d$~$=$~$50$ variables, where each local conditional distribution is parameterized by a 2-layer neural network with five hidden nodes.
    In this setting, the total number of conditional distribution parameters in a given BN amounts to $|\Thetab|$~$=$~13,050 weights and biases.
    The metrics are aggregated for inference of 30 random BNs of each graph type.
    Here, DiBS only infers 10 particles to make the wall time on CPUs comparable to M-\mcmcmc and G-\mcmcmc.
    As for posterior inference of BNs with $d$~$=$~$20$ variables, DiBS and DiBS+ perform favorably compared to the \mcmcmc baselines.
    \vspacefigurecaption 
    }
    \label{fig:fcgauss50}
\end{figure}

\section{Additional Analyses and Ablation Studies}\label{app:results-additional}
Having compared DiBS with several alternative approaches to Bayesian structure learning in Section \ref{sec:evaluation}, this supplementary section is devoted to a more in depth analysis of some of its properties. 
This is done by changing, or leaving out single design aspects of the algorithm and studying the effect on the previous metrics.

As in Section \ref{sec:evaluation}, DiBS and its instantiation with SVGD are used interchangeably here, and DiBS+ denotes the weighted mixture of particles.
Since the metrics do not qualitatively differ between inference of \erdosrenyi and scale-free BN structures in our experiments of Section \ref{sec:evaluation}, we only consider the former here.
Unless mentioned otherwise, the following experimental setup corresponds to \emph{joint} posterior inference of linear Gaussian BNs with $d$~$=$~$20$ variables in Section \ref{ssec:results-lingauss}.

\subsection{Graph Embedding Representation}

\begin{figure}[!b]
    \centering
    \hspace*{-10pt}
    \includegraphics{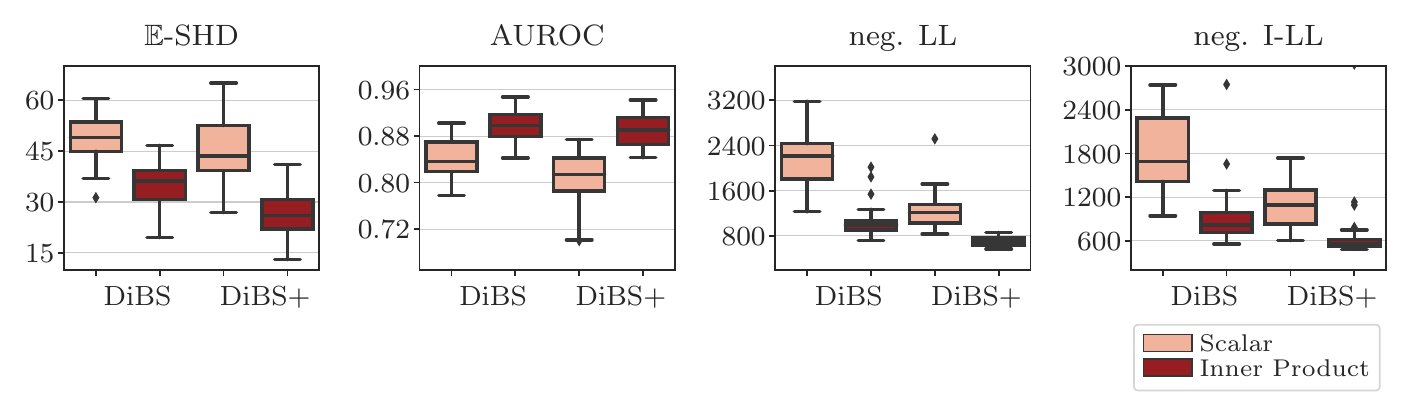} 
    \vspace*{-12pt} 
    \caption{
    Contrasting the bilinear graph model of Section \ref{sec:graph-latent-space} with its more trivial variant, where each latent variable models the edge probabilities directly via the sigmoid.
    The plots aggregate the results for joint inference of 30 randomly generated linear Gaussian BNs with $d$~$=$~$20$ variables.
    }
    \vspace*{-5pt} 
    \label{fig:ablation-embed}
\end{figure}
In Section \ref{sec:graph-latent-space}, we propose to use a generative graph model $p_\alpha(\Gb \given \Zb)$ that is based on the inner product of latent embeddings for each node.
In particular, we choose $p_\alpha(g_{ij} = 1 \given \Zb) = \sigma_\alpha(\ub^\top_i \vb_j)$ with latent variables $\Zb = [\Ub, \Vb]$.
In Figure \ref{fig:ablation-embed}, we contrast this modeling choice with the more trivial variant
$p_\alpha(g_{ij} = 1 \given \Zb) = \sigma_{\alpha}(\zb_{ij})$, where single \emph{scalars} rather than inner products between latent vectors encode the edge probabilities.
\citet{bengio2020meta} and \citet{ke2019learning} use the scalar variant with fixed $\alpha$~$=$~$1$ in the context of causal inference.

The comparison in Figure \ref{fig:ablation-embed} illustrates that incorporating only the bilinear parameterization of edge probabilities in the generative graph model improves performance by a significant margin. 
We hypothesize that the coupling between edges results in smoother densities, which might be less prone to local minima in gradient-based methods such as DiBS.

\begin{figure}[!t]
    \centering
    \hspace*{-20pt}
    \includegraphics{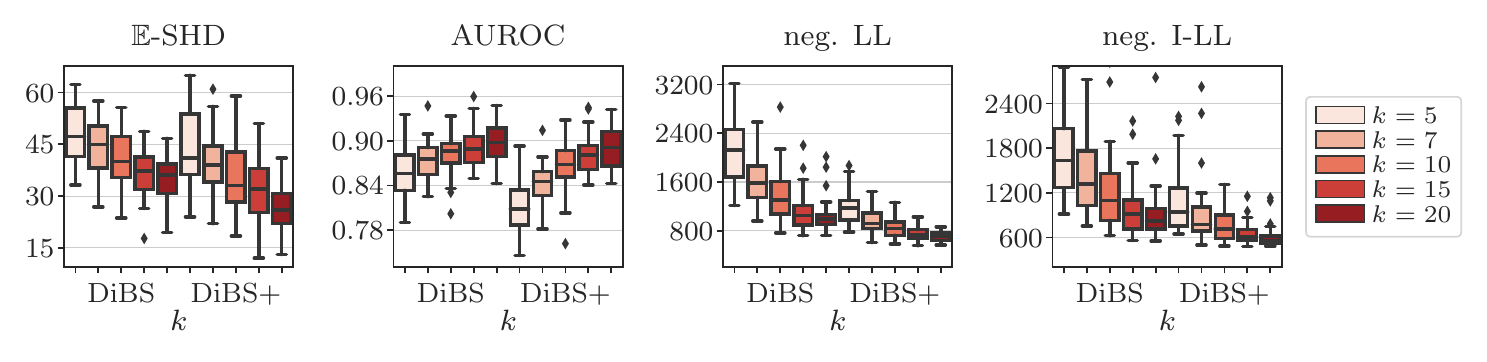} 
    \caption{
    DiBS for joint inference of linear Gaussian BNs with $d$~$=$~$20$ variables for different sizes of the latent variables $\Zb \in \RR^{2 \times d \times k}$.
    Lower rank parameterizations of the matrix of edge probabilities balance the tradeoff between computational efficiency and posterior approximation quality. 
    }
    \label{fig:ablation-dim}
\end{figure}

\subsection{Graph Embedding Dimensionality}
Another feature of the inner product representation of graphs is the ability to control the dimensionality of the posterior inference task.
As described in Section \ref{sec:dibs-svgd}, we generally set $k$~$=$~$d$ for the latent variables $\Zb \in \RR^{2 \times d \times k}$ that parameterize our graph model $p_\alpha(\Gb \given \Zb)$.
This leaves the matrix of edge probabilities fully expressible and without a rank constraint.
In principle, however, the formulation in (\ref{eq:graph-latent-model}) allows us to arbitrarily vary $k$.
This creates a trade-off between the complexity of the parameterization and the tractability and dimensionality challenges in approximate inference of $p(\Zb \given \Dcal)$ or $p(\Zb, \Thetab \given \Dcal)$.
Limiting $k$~$<$~$d$ has connections to the theory of low-rank realizations of sign matrices.

\begin{figure}[!b]
    \centering
    \hspace*{-22pt}
    \includegraphics{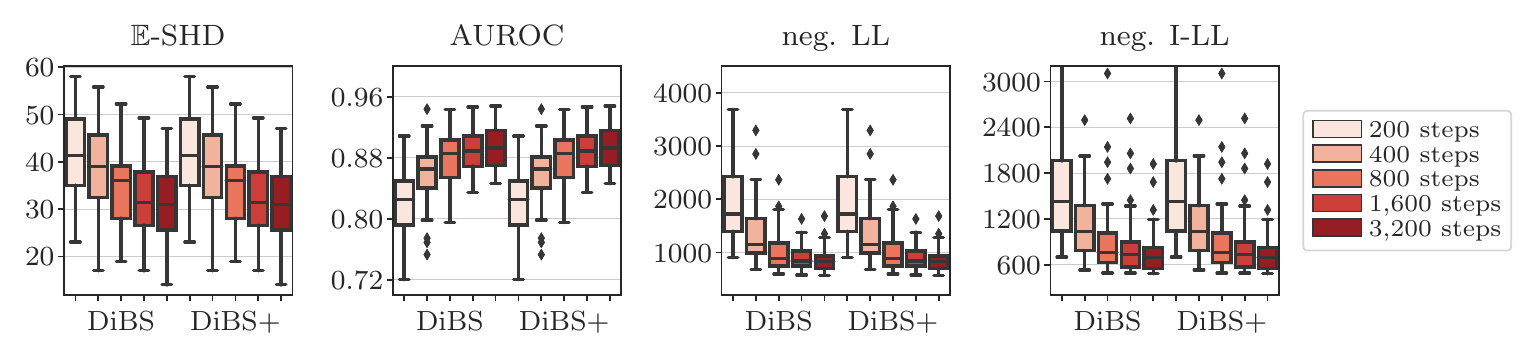} 
    \caption{
    Performance of DiBS and DiBS+ as a function the number of particle transport steps $T$.
    As previously, the plots aggregate the results for inference of 30 randomly generated linear 20-node Gaussian BNs.
    The latent variable $\Zb$ is specified with its default dimensions $k$~$=$~$d$~$=$~$20$.
    After already roughly 1,000 iterations, DiBS and DiBS+ obtain good posterior approximations.
    }
    \label{fig:ablation-steps}
\end{figure}

We perform inference with DiBS for $k \in \{5, 7, 10, 15, 20\}$, leaving all other aspects of the algorithm unchanged.
Hence, the corresponding posterior over $\Zb$ has $\{200, 280, 400, 600, 800\}$ dimensions, respectively.
The results in Figure \ref{fig:ablation-dim} suggest that lower values of $k=15$, or even $k=10$, are already able to achieve competitive performance across all metrics.
Interestingly, the structural $\EE$-SHD metric appears to suffer most from a small loss in complexity.

In this context, one should keep in mind that the bandwidth parameters $\gamma_z$ and $\gamma_\theta$ were set to achieve good performance with $k$~$=$~$d$~$=$~$20$.
It is possible that lower values of $k$ can reach performances that are even closer to the full-rank variant of DiBS with alternative settings for $\gamma_z$ and $\gamma_\theta$.
In addition, a lower-rank DiBS variant could be particularly promising for inference of very large BNs, where the computational challenges of the full $O(d^2)$ latent representation might outweigh its benefits in terms of expressibility.

\subsection{Particle Transport Iterations}
Since DiBS uses Stein variational gradient descent~\citep{liu2016stein} for posterior inference, our method iteratively transports a set of latent graph particles, or latent graph and parameter tuples, for a number of $T$ steps.
As the particles are randomly initialized, the approximation quality of SVGD particles (and thus also DiBS particles) improves with the number of steps. 
We are interested in the degree to which a small number of transport steps provide a good posterior approximation in the face of computational constraints.

In Figure \ref{fig:ablation-steps}, we show the performance of DiBS as a function of the number of transport steps.
We find that even a smaller number of iterations achieves competitive results across the metrics.
In addition, the variance of performance in the predictive metrics neg.\ LL and neg.\ I-LL decreases monotonically as a function of the performed transforms, whereas the variation in $\EE$-SHD remains roughly the same.

\subsection{Uncertainty Quantification Within a Markov Equivalence Class}
%
%
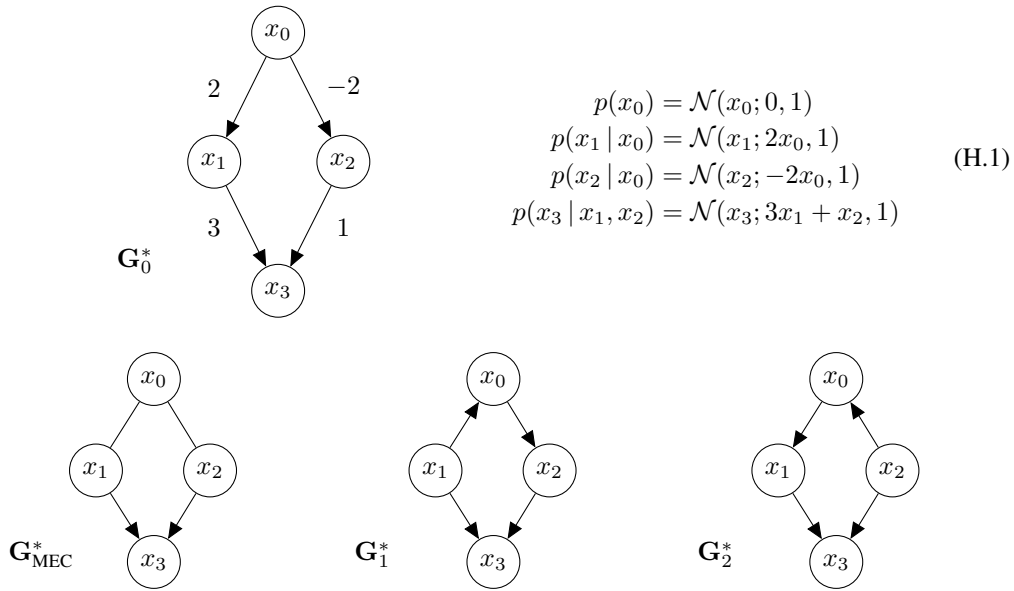
\begin{figure}[!b]
\begin{minipage}{.55\textwidth}
\centering
\begin{tikzpicture}

  \node[latent] (x1) {$x_1$};
  \node[latent, right=of x1] (x2) {$x_2$};
  \node[latent, above=of x1, xshift=0.85cm] (x0) {$x_0$};
  \node[latent, below=of x1, xshift=0.85cm] (x3) {$x_3$};
  \node[const, left=of x3, xshift=-0.3cm, yshift=0.4cm]  (label) {$\Gb^*_0$};

  \edge {x0} {x1} ; %
  \edge {x0} {x2} ; %
  \edge {x1} {x3} ; %
  \edge {x2} {x3} ; %
  
  \node[const, above=of x1, yshift=-0.5cm, xshift=0.0cm] (label01) {$2$};
  \node[const, below=of x1, yshift=0.6cm, xshift=0.0cm] (label13) {$3$};
  \node[const, above=of x2, yshift=-0.5cm, xshift=0.0cm] (label02) {$-2$};
  \node[const, below=of x2, yshift=0.6cm, xshift=0.0cm] (label23) {$1$};

\end{tikzpicture}
\end{minipage}
\begin{minipage}{.45\textwidth}
\vspace*{-10pt}
\begin{align}
\hspace*{-30pt}
\begin{split}
	p(x_0) &= \Ncal(x_0; 0, 1)\\
	p(x_1 \given x_0) &= \Ncal(x_1; 2x_0, 1)\\
	p(x_2 \given x_0) &= \Ncal(x_2; -2x_0, 1)\\
	p(x_3 \given x_1, x_2) &= \Ncal(x_3; 3x_1 + x_2, 1)
\end{split}
\end{align}
\end{minipage}

\bigskip

\begin{minipage}{.32\textwidth}
\centering
\begin{tikzpicture}

  \node[latent] (x1) {$x_1$};
  \node[latent, right=of x1, xshift=-0.2cm] (x2) {$x_2$};
  \node[latent, above=of x1, xshift=0.77cm, yshift=-0.5cm] (x0) {$x_0$};
  \node[latent, below=of x1, xshift=0.77cm, yshift=0.5cm] (x3) {$x_3$};
  \node[const, left=of x3, xshift=0.3cm, yshift=0.1cm]  (label) {$\Gb^*_{\text{MEC}}$};

  \factoredge {x0} {x1} {} ; %
  \factoredge {x0} {x2} {} ; %
  \edge {x1} {x3} ; %
  \edge {x2} {x3} ; %
  
\end{tikzpicture}
\end{minipage}
\begin{minipage}{.32\textwidth}
\centering
\begin{tikzpicture}

  \node[latent] (x1) {$x_1$};
  \node[latent, right=of x1, xshift=-0.2cm] (x2) {$x_2$};
  \node[latent, above=of x1, xshift=0.77cm, yshift=-0.5cm] (x0) {$x_0$};
  \node[latent, below=of x1, xshift=0.77cm, yshift=0.5cm] (x3) {$x_3$};
  \node[const, left=of x3, xshift=-0.0cm, yshift=0.1cm]  (label) {$\Gb^*_1$};

  \edge {x1} {x0} {} ; %
  \edge {x0} {x2} {} ; %
  \edge {x1} {x3} ; %
  \edge {x2} {x3} ; %
  
\end{tikzpicture}
\end{minipage}
\begin{minipage}{.32\textwidth}
\centering
\begin{tikzpicture}

  \node[latent] (x1) {$x_1$};
  \node[latent, right=of x1, xshift=-0.2cm] (x2) {$x_2$};
  \node[latent, above=of x1, xshift=0.77cm, yshift=-0.5cm] (x0) {$x_0$};
  \node[latent, below=of x1, xshift=0.77cm, yshift=0.5cm] (x3) {$x_3$};
  \node[const, left=of x3, xshift=-0.0cm, yshift=0.1cm]  (label) {$\Gb^*_2$};

  \edge {x0} {x1} ; %
  \edge {x2} {x0} ; %
  \edge {x1} {x3} ; %
  \edge {x2} {x3} ; %
  
\end{tikzpicture}
\end{minipage}
\caption{
Four-node example linear Gaussian Bayesian network. Under the BGe marginal likelihood and a uniform prior, $\Gb^*_{0}$, $\Gb^*_{1}$, and $\Gb^*_{2}$ are scored equally.
While the v-structure $x_1 \rightarrow x_3 \leftarrow x_2$ is an identifiable feature of the MEC of $\Gb_0^*$ and thus present in $\Gb_{\text{MEC}}^*$, the edge directions of $x_1$\,---\,$x_0$\,---\,$x_2$ cannot be distinguished even given infinite observational data.
}
\label{fig:mec-toy}
\end{figure}
When performing posterior inference of $p(\Gb \given \Dcal)$ with the BGe marginal likelihood and a uniform prior $p(\Gb)$, each Markov equivalent structure is assigned equal likelihood.
This might be desirable considering that causal edge directions are often not fully identifiable from purely observational data.
As DiBS infers a posterior over DAGs rather than MECs, we aim to validate the ability of DiBS to correctly quantify the uncertainty present in nonidentifiable edge directions.

To this end, we consider a 4-node example Bayesian network, small enough to allow for the closed-form computation of the ground-truth posterior by exhaustive enumeration of all possible DAGs.
This enables us to compute the true single and pairwise posterior edge marginals and contrast them with the approximate posterior marginals inferred by DiBS. 
The graph structure for this analysis is chosen to contain both an identifiable v-structure and a nonidentifiable edge pair.
Figure \ref{fig:mec-toy} shows the ground truth DAG $\Gb^*_0$, its linear Gaussian parameters, and the observational model.
In addition, Figure \ref{fig:mec-toy} lists the essential graph $\Gb^*_{\text{MEC}}$ as well as the two other Markov equivalent DAGs $\Gb^*_1$ and $\Gb^*_2$ in the MEC represented by $\Gb^*_{\text{MEC}}$.

We perform marginal posterior inference with DiBS using the experimental setup and hyperparameters for 20-node linear Gaussian BNs.
In this example setting, DiBS employs a uniform prior over graphs and uses the default $k$~$=$~$d$~$=$~$4$.
Table \ref{tab:results-mec-toy} shows the ground truth and inferred pairwise edge marginals under the posterior.
We find that DiBS+ correctly infers both the uncertainty in the edge directions of $x_1$\,---\,$x_0$\,---\,$x_2$ as well as the high confidence in the presence of the v-structure $x_1 \rightarrow x_3 \leftarrow x_2$.
While the unweighted particles of DiBS do not exhibit false confidence in structures that are not present in $\Gb_0^*$, its inferred degree of uncertainty is too high compared to the ground truth.
The DiBS+ variant overcomes the inexact empirical average of DiBS by weighting the particles by their unnormalized posterior probabilities.

\begin{table}[!t]
\caption{
Ground truth and average inferred posterior marginals given $N=100$ observations from the ground truth model in Figure \ref{fig:mec-toy}.
Listed are the probabilities for the nonidentifiable edge structure $x_1$\,---\,$x_0$\,---\,$x_2$  (top) and the identifiable v-structure $x_1 \rightarrow x_3 \leftarrow x_2$ (bottom).
Averaged over 30 random particle initilizations, DiBS+ correctly quantifies the confidence and uncertainty in the v-structure and nonidentifable edge pair, respectively.
}\label{tab:results-mec-toy}
\centering
\vspace{5pt}
\begin{tabular}{c c c c}
	& DiBS & DiBS+ & \textbf{Ground Truth} \\\hline
	$p(x_1 \rightarrow x_0, x_0 \rightarrow x_2 \given \Dcal)$ 
	& 0.134 & 0.216 & \textbf{0.298} \\	
	$p(x_1 \rightarrow x_0, x_0 \leftarrow x_2 \given \Dcal)$ 
	& 0.201 & 0.037 & \textbf{0.052} \\	
	$p(x_1 \leftarrow x_0, x_0 \rightarrow x_2 \given \Dcal)$ 
	& 0.223 & 0.409 & \textbf{0.311} \\	
	$p(x_1 \leftarrow x_0, x_0 \leftarrow x_2 \given \Dcal)$ 
	& 0.103 & 0.298 & \textbf{0.297} \\	\hline
	$p(x_1 \rightarrow x_3, x_3 \rightarrow x_2 \given \Dcal)$ 
	& 0.157 & 0.013 & \textbf{0.017} \\	
	$p(x_1 \rightarrow x_3, x_3 \leftarrow x_2 \given \Dcal)$ 
	& 0.338 & 0.934 & \textbf{0.914} \\		
	$p(x_1 \leftarrow x_3, x_3 \rightarrow x_2 \given \Dcal)$ 
	& 0.275 & 0.034 & \textbf{0.043} \\	
	$p(x_1 \leftarrow x_3, x_3 \leftarrow x_2 \given \Dcal)$ 
	& 0.154 & 0.019 & \textbf{0.025} \\	\hline
\end{tabular}
\centering
\end{table}

\section{Experimental Details for Application to Protein Signaling Networks}\label{app:application-details}
\looseness - 1 The data by \citet{sachs2005causal} as well as the corresponding consensus graph used in Section \ref{sec:application} are taken as provided by the \texttt{CausalDiscoveryToolbox} \citep{kalainathan2019causal}, which is published under an MIT Licence.
We standardize the data for inference.
Because $N$ is large, DiBS uses minibatches of 100 observations to estimate the scores of the posterior.
All hyperparameters and BN specifications are chosen by default exactly as during the synthetic evaluation of linear and nonlinear Gaussian BNs in Table \ref{tab:dibs-params}, respectively, except that DiBS correspondingly uses $k$~$=$~$d$~$=$~$11$.
For joint posterior inference of linear Gaussian BNs, BPC${}^{*}$ and BGES${}^{*}$ still use the BGe marginal likelihood; since metrics are very similar to BPC and BGES, their scores are not reported.

In line with inference on synthetic data in Section \ref{sec:evaluation}, 
the BGe marginal likelihood employed for the experiments in Section \ref{sec:application} uses the default effective sample sizes $\alpha_\mu = 1$ and $\alpha_{\omega}= d+2$ described in Appendix \ref{app:gaussian-bns}.
Likewise, we again set the noise level for inference with the explicitly parameterized linear and nonlinear Gaussian networks to $\sigma^2$~$=$~$0.1$. 

Since the effective sample size $\alpha_\mu$ and the noise level $\sigma^2$ may affect the model complexity of the inferred BNs, 
\eg, the mean number of inferred edges in the DAG, we provide additional results for alternative values of these Bayesian network model hyperparameters in Tables \ref{tab:results-sachs-additional-bge} and \ref{tab:results-sachs-additional-joint}.
Overall, we find that increasing the effective sample size $\alpha_\mu$ does not significantly change metrics across the considered methods. However, higher fixed noise levels $\sigma^2$ do result in less inferred edges, which tends to lead to lower $\EE$-SHD but worse AUROC, \ie, less calibrated edge confidence scores. 
We note that these are not free parameters of the inference methods that approximate the posterior, but specifications of the inferred BN models themselves.

\newpage

\begin{table}[!ht]
\caption{
Additional results for marginal posterior inference of protein signaling pathways under the BGe marginal likelihood of linear Gaussian BNs~\citep{geiger1994learning,geiger2002parameter}.
Changing the effective sample size in the BGe Normal-Wishart prior does not result in significantly different metrics compared to the default $\alpha_\mu = 1$ used in all of our experiments.
For $\alpha_\mu = 10$, DiBS and DiBS+ average an expected number of 39.6 and 35.4 edges, respectively.
Metrics are the mean $\pm$ SD of 30 random restarts. 
}\label{tab:results-sachs-additional-bge}
\centering
\vspace{5pt}
\begin{tabular}{l c c}
& \multicolumn{2}{c}{$\alpha_\mu = 10$} \\
\cline{2-3}
&  $\EE$-\textbf{SHD} & \textbf{AUROC}  \\
\cline{1-3}
\mcmcmc        & 34.3 $\pm$ 0.4  &  0.622 $\pm$ 0.020 \\
BPC            & 25.5 $\pm$ 2.3  &  0.566 $\pm$ 0.020 \\
BGES           & 33.8 $\pm$ 1.8  &  0.641 $\pm$ 0.034 \\
DiBS           & 37.9 $\pm$ 0.5  &  0.637 $\pm$ 0.046 \\
DiBS+          & 35.1 $\pm$ 1.8  &  0.627 $\pm$ 0.050 \\
\cline{1-3}
\end{tabular}
\centering
\end{table}

\begin{table}[!ht]
\caption{
Additional results for joint posterior inference of protein signaling pathways under explicitly parameterized linear (top) and nonlinear (bottom) Gaussian BNs.
The hyperparameter $\sigma^2$ specifies the noise level underlying the inferred Bayesian networks.
We find that higher noise levels $\sigma^2$ tend to result in less edges. 
When inferring linear Gaussian BNs with $\sigma^2=0.01$ ($\sigma^2 = 1$), DiBS averages an expected number of 11.6 (8.8) edges, DiBS+ 13.8 (9.9) edges.
For nonlinear Gaussian BNs with $\sigma^2=0.01$ ($\sigma^2 = 1$), DiBS averages 15.6 (5.2) edges, DiBS+ 17.5 (6.8) edges.
Due to less false positives, the $\EE$-SHD improves, but the degree of uncertainty in the presence of edges is quantified less accurately, resulting in worse AUROC.
Metrics are the mean $\pm$ SD of 30 random restarts. 
}\label{tab:results-sachs-additional-joint}
\centering
\vspace{5pt}
\begin{tabular}{l c c c c}
& \multicolumn{2}{c}{$\sigma^2 = 0.01$} & \multicolumn{2}{c}{$\sigma^2 = 1$} \\
\cline{2-5}
&  $\EE$-\textbf{SHD} & \textbf{AUROC} &  $\EE$-\textbf{SHD} & \textbf{AUROC} \\
\cline{1-5}
M-\mcmcmc        & 38.1 $\pm$ 3.4  &  0.536 $\pm$ 0.082 & 33.1 $\pm$ 3.4  &  0.543 $\pm$ 0.105  \\
G-\mcmcmc        & 30.9 $\pm$ 3.0  &  0.518 $\pm$ 0.051 & 29.8 $\pm$ 3.7  &  0.531 $\pm$ 0.078 \\
DiBS             & 23.0 $\pm$ 0.5  &  0.595 $\pm$ 0.069 & 20.3 $\pm$ 0.4  &  0.601 $\pm$ 0.039  \\
DiBS+            & 22.9 $\pm$ 2.0  &  0.540 $\pm$ 0.048 & 20.0 $\pm$ 1.4  &  0.569 $\pm$ 0.040 \\
\cline{1-5}
\end{tabular}
\medskip

\begin{tabular}{l c c c c}
& \multicolumn{2}{c}{$\sigma^2 = 0.01$} & \multicolumn{2}{c}{$\sigma^2 = 1$} \\
\cline{2-5}
&  $\EE$-\textbf{SHD} & \textbf{AUROC} &  $\EE$-\textbf{SHD} & \textbf{AUROC} \\
\cline{1-5}
M-\mcmcmc        & 38.7 $\pm$ 3.2  &  0.555 $\pm$ 0.101  &  18.4 $\pm$ 0.1  &  0.501 $\pm$ 0.043 \\
G-\mcmcmc        & 34.9 $\pm$ 3.6  &  0.542 $\pm$ 0.064  &  30.6 $\pm$ 2.5  &  0.538 $\pm$ 0.059 \\
DiBS             & 24.3 $\pm$ 0.6  &  0.582 $\pm$ 0.050  &  17.7 $\pm$ 0.1  &  0.550 $\pm$ 0.020  \\
DiBS+            & 24.9 $\pm$ 2.9  &  0.535 $\pm$ 0.045  &  18.5 $\pm$ 0.5  &  0.530 $\pm$ 0.028\\
\cline{1-5}
\end{tabular}
\centering
\end{table}

\vfill

\end{document}